\documentclass[times,10pt,twocolumn, letterpaper]{article}
\usepackage[final]{cvpr}
\usepackage{times}

\newcommand{\todo}[1]{{\color{red}#1}}

\usepackage{amsmath,amsfonts,bm}

\def\eqref#1{equation~\ref{#1}}

\def\1{\bm{1}}

\DeclareMathAlphabet{\mathsfit}{\encodingdefault}{\sfdefault}{m}{sl}
\SetMathAlphabet{\mathsfit}{bold}{\encodingdefault}{\sfdefault}{bx}{n}

\usepackage{hyperref}
\usepackage{url}
\usepackage{xspace}
\usepackage{booktabs}
\usepackage{amssymb}
\usepackage{pifont}
\usepackage{graphicx}
\usepackage{amsmath}
\usepackage{pifont}
\usepackage{multirow}
\usepackage{mathtools}
\usepackage{array}  
\usepackage{arydshln}
\usepackage{tabularx}
\usepackage{soul}

\usepackage[accsupp]{axessibility}  

\newcommand{\codename}[0]{\textsc{BHCast}\xspace}
\newcommand{\M}[0]{$\,GMc^{-3}$\,}
\newcommand{\mycomment}[1]{}

\def\paperID{34417} 
\def\confName{CVPR}
\def\confYear{2026}

\title{\codename: Unlocking Black Hole Plasma Dynamics from a Single Blurry Image with Long-Term Forecasting}

\author{
    Renbo Tu$^{1,2}$ \quad 
    Ali SaraerToosi$^{1,2}$ \quad 
    Nicholas S. Conroy$^3$ \quad 
    Gennady Pekhimenko$^{1,2,4}$ \quad 
    Aviad Levis$^{1,2}$ \\
    $^1$\textit{University of Toronto} \quad $^2$ \textit{Vector Institute} \quad
    $^3$ \textit{University of Illinois at Urbana-Champaign} \quad $^4$ \textit{NVIDIA} \\
}

\begin{document}

\maketitle

\begin{abstract}
The Event Horizon Telescope (EHT) delivered the first image of a black hole by capturing the light from its surrounding accretion flow, revealing structure but not dynamics. Simulations of black hole accretion dynamics are essential for interpreting EHT images but costly to generate and impractical for inference. Motivated by this bottleneck, \codename presents a framework for forecasting black hole plasma dynamics from a single, blurry snapshot, such as those captured by the EHT. At its core, \codename is a neural model that transforms a static image into forecasted future frames, revealing the underlying dynamics hidden within one snapshot. With a multi-scale pyramid loss, we demonstrate how autoregressive forecasting can simultaneously super-resolve and evolve a blurry frame into a coherent, high-resolution movie that remains stable over long time horizons. From forecasted dynamics, we can then extract interpretable spatio-temporal features, such as pattern speed (rotation rate) and pitch angle. Finally, \codename uses gradient-boosting trees to recover black hole properties from these plasma features, including the spin and viewing inclination angle. The separation between forecasting and inference provides modular flexibility, interpretability, and robust uncertainty quantification. We demonstrate the effectiveness of \codename on simulations of two distinct black hole accretion systems, Sagittarius A* and M87*, by testing on simulated frames blurred to EHT resolution and real EHT images of M87*. Ultimately, our methodology establishes a scalable paradigm for solving inverse problems, demonstrating the potential of learned dynamics to unlock insights from resolution-limited scientific data.

\end{abstract}
\vspace{-10pt}
\section{Introduction} \vspace{-5pt}
\label{sec:intro}
\begin{figure}
    \centering
    \includegraphics[width=1.0\linewidth]{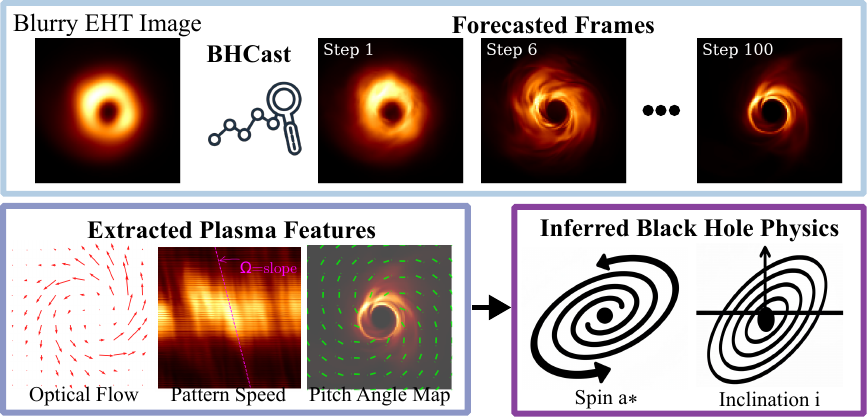}
    \captionsetup{belowskip=0pt}
    \caption{\textbf{From a Single Blurry Frame to Black-Hole Dynamics.} A single blurry image, reconstructed from Event Horizon Telescope measurements, is used to forecast a dynamic sequence of high-resolution frames. From this forecasted movie, features of accretion disk plasma can be extracted to infer the black hole's physical properties. This sequential framework—forecasting, feature extraction, and parameter estimation—is robust, versatile, and fully interpretable, providing a new pathway for analyzing horizon-scale dynamics.}\label{fig:sgra_visual}
\end{figure}
The Event Horizon Telescope (EHT) has captured the first images of M87* and Sagittarius A* (Sgr A*) black holes, a landmark in both astrophysics and imaging science, achieving the highest angular resolution ever recorded in astronomy~\cite{m87_paper1, sgr_a_paper1}. To reach this resolution, data from worldwide radio telescopes are combined using Very Long Baseline Interferometry (VLBI). The result is the first visual evidence of a black hole event horizon. Nonetheless, these static images provide limited insight into the complex plasma dynamics that power black hole accretion. Capturing and understanding their temporal evolution is key to interpreting EHT data and probing the underlying plasma dynamics~\cite{conroy2023rotation}. 

\vspace{-2pt}

To bridge the gap between static images and dynamics, astrophysicists are developing extensive libraries of time-dependent General Relativistic Magnetohydrodynamics (GRMHD) simulations \cite{ipol_paper, kharma_prather_2024} which involve solving nonlinear equations of general relativity coupled to multi-scale magneto-hydrodynamics equations for magnetized fluids \cite{etienne2015illinoisgrmhd}. These highly-complex simulations, for the most part, show remarkable consistency with observation data \cite{akiyama2019event, akiyama2022first, wong_patoka_2022}. Inferring black hole properties from such simulations relies on scanning a massive library of GRMHD simulations to find the best match to observations. This process is computationally prohibitive, as a single simulation can take weeks on a supercomputer, making exploration of large or fine-grained parameter spaces infeasible~\cite{etienne2015illinoisgrmhd}.

This computational bottleneck motivates our work (Fig.~\ref{fig:sgra_visual}), which builds on recent advances in physics-informed deep learning~\cite{raissi2019physics, karniadakis2021physics}. We replace the expensive forward GRMHD simulation with a lightweight neural surrogate, which is highly efficient on modern GPUs. Neural surrogate models for fluid dynamics have gained significant traction in recent years, but extending them to black hole plasmas from reconstructed (rather than simulated) EHT images introduces a unique set of challenges ~\cite{li2020fourier, li2021learning, raissi2019physics, chen2018neural}. First, EHT images are fundamentally resolution-limited by the telescope's $\sim20 \mu as$ (micro-arcsecond) beam, creating an ill-posed inverse problem \cite{akiyama2022first}. Recovering the lost high-frequency information, a super-resolution task, is essential to reveal the subtle cues of plasma dynamics encoded in these images. Second, key indicators of black hole dynamics require stable forecasts over long temporal horizons of 300 $GMc^{-3}$\footnote{For gravitational constant G, black hole mass M, and speed of light c, $GMc^{-3}$ is the characteristic dynamical timescale near the event horizon which depends on the black hole mass: ${\sim}20$ seconds for Sgr~A* and ${\sim}9$ hours for M87*.} to be measured accurately \cite{conroy2023rotation}. Existing research focuses on single-step predictions, while autoregressive forecasts of long-term dynamics remain a major challenge due to rapid error accumulation~\cite{raissi2019physics,li2020fourier, li2021learning}. 

\begin{figure}
    \centering
    \includegraphics[width=\linewidth]{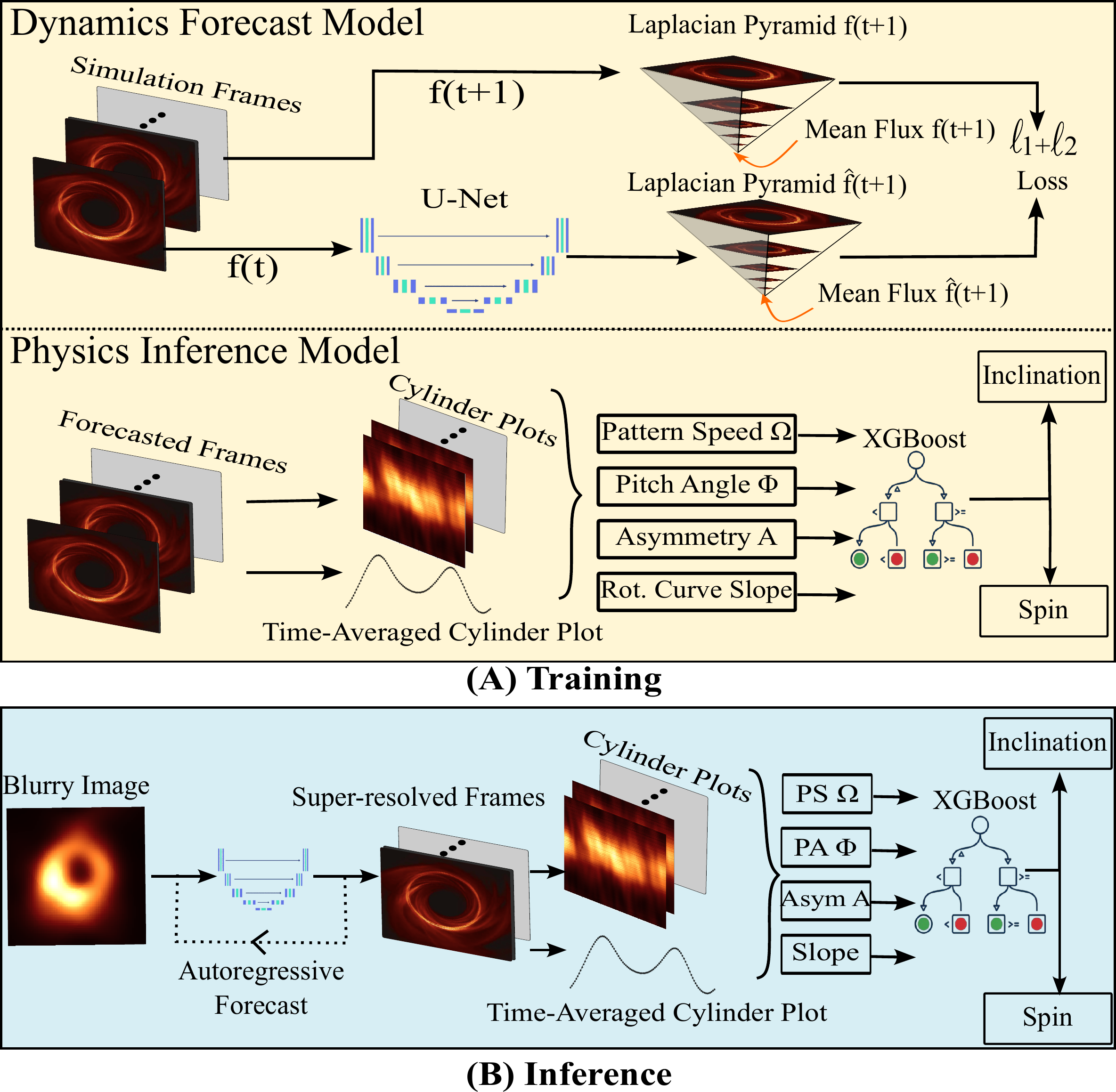}
    
    \caption{
    \textbf{Training and inference pipelines of \codename.} We train the dynamics forecast model (U-Net) to predict the next frame and the physics inference model (XGBoost) to estimate black hole parameters. During inference, we autoregressively forecast with the U-Net to produce cylinder plots to extract plasma features, which are inputs to XGBoost for physics inference. } 
    \label{fig: overview}
\end{figure}

In this paper, we introduce \codename, a long-term forecasting framework designed to produce high-fidelity, super-resolved simulation dynamics from a single, blurry black hole image. Our framework overcomes the instability of autoregressive prediction by incorporating a multi-scale pyramid loss during training. We demonstrate this super-resolution effect and the stability of our predictions in simulations (Fig.~\ref{fig: overview}). Using the forecasted frames, we extract key plasma features that characterize the dynamic and geometric properties of the black hole accretion flow. As an application of \codename for inverse problems, we train an XGBoost model to predict black hole parameters from the computed plasma features. Note that our framework provides interpretable inference of plasma and black hole parameters, as recovered parameters can be traced back to visual cues in the forecasted frames. Furthermore, our modular design allows for future improvements across different stages of the \codename pipeline.

We evaluate the framework in forecasting, feature extraction, and parameter inference. For long-term forecasting, we demonstrate that \codename produces super-resolved and temporally stable predictions, outperforming both end-to-end deep learning and traditional computer vision baselines. For plasma feature extraction and black hole parameter inference, \codename is more robust to noise compared to direct inference with a black-box supervised baseline.

In summary, we make the following core contributions:
\begin{enumerate}
	\item We recast a challenging, ill-posed astrophysical imaging task as a forecasting+inference problem, enabling the use of efficient neural surrogates for rapid black hole characterization. Our reformulation provides a useful roadmap for future works in scientific imaging. 
	\item Leveraging a physics-informed multi-scale loss, our dynamics surrogate model super-resolves blurry input images and forecasts stable evolution for up to 100 steps, (500 \M or 10,000 seconds on Sgr A*), sufficiently long for precise physical measurements. We demonstrate forecasting on real EHT images of M87*
	\item \codename produces high-fidelity forecasts with accurate dynamic and geometric plasma features. On four features, the forecast error is comparable with a baseline of four ResNet models trained separately on each feature. 
	\item For parameter inference, \codename estimates black hole spin and inclination angle better than ResNet, demonstrating better robustness to input noise compared to direct inference.
\end{enumerate}

\section{Related Works}
\label{sec:related}
\paragraph{Multi-Scale and Self-Similarity in Vision}
Multi-scale representations are foundational in computer vision, from classical pyramid decompositions~\cite{burt1987laplacian} to coarse-to-fine inference and internal image statistics. Natural images exhibit strong cross-scale self-similarity, a principle exploited in seminal works on super-resolution and deblurring~\cite{levin2009understanding, michaeli2013nonparametric, glasner2009super, zontak2011internal}. These methods show that fine-scale structure can often be reconstructed from recurring patterns already present within a single image. 
Although developed for natural images, this multi-scale and self-similar perspective aligns well with the fractal, turbulent structure of black hole plasmas, motivating our use of a hierarchical loss to recover high-frequency detail from sparse, resolution-limited data.

\paragraph{Modeling Dynamics} \vspace{-10pt}
Neural surrogates for dynamical systems typically follow one of two strategies: next-frame prediction or full-sequence generation. Data-hungry diffusion-based approaches for GRMHD data fall into the latter category, generating entire video sequences in a single pass but requiring the forecast horizon to be fixed a priori ~\cite{zhang2025step}. Next-frame prediction is more versatile, as a model can be rolled out autoregressively for arbitrarily long horizons, but this flexibility comes at the cost of compounding errors and potential instability during rollout~\cite{song2025history}. While Fourier Neural Operators are strong in learning PDE-driven dynamics, they are better suited for systems with global behaviors rather than black holes with local turbulence ~\cite{li2020fourier, li2021learning}. For these reasons, \codename adopts a data-efficient U-Net for next-frame prediction and explicitly addresses long-horizon stability to recover plasma evolution.


\paragraph{Black Hole Parameter Estimation} \vspace{-10pt}
Inferring physical parameters from EHT observations typically follows one of two approaches: direct inference from EHT visibilities in the frequency domain, or image-based inference in the pixel domain.
Recent visibility-domain methods use adaptive Bayesian inference, but their non-parallelizable MCMC loops create a major computational bottleneck, with individual runs taking up to 20 CPU hours~\cite{yfantis2024testing}. To avoid this cost, feed-forward machine learning approaches based on visibility data products have been used for classification~\cite{thyagarajan2024interferometric}, however, this approach is limited to inference of simple physical models. Conversely, image-based inference using supervised learning, explored by Deep Horizon and its extensions, offers a faster alternative~\cite{van2020deep, farah2025machine}. The primary drawback of these "black-box" approaches is their failure to capture black hole dynamics, limiting the physical interpretability of their predictions. Other methods, such as those based on random fields, can infer from both visibilities and images, but they have thus far been limited to a small, fixed set of parameters~\cite{levis2021inference}.


\section{Background}
\label{sec:background}

\paragraph{Black Hole EHT-Imaging}
The EHT achieves an unprecedented angular resolution of $\sim20\,\mu$as by combining millimeter and sub-millimeter radio telescope observations across the globe~\cite{sgr_a_paper1}. Despite the resolution, reconstructed images remain inherently uncertain due to various factors including interstellar scattering, sparse coverage, and source variability \citep{Johnson_2018, Issaoun_2021, sgr_a_paper1, sgr_a_paper3}. Together, these effects imply that EHT images can be modeled as a convolution of the intrinsic black hole emission structure with both astrophysical and instrumental blurring. Despite these challenges, our hypothesis is that the EHT images contain rich dynamical information encoded in the blurry image morphology. Two quantities proposed to characterize black hole dynamics are: (i) the \textbf{pitch angle}, which is the angle between a spiral pattern and the local azimuthal direction \cite{Ricarte_2022_pitch_angle, baubock_spiral_prep}, and (ii) the \textbf{pattern speed}, which is the angular rate at which a coherent structure (like a spiral arm or brightness pattern) rotates~\cite{conroy2023rotation}.

\paragraph{GRMHD Simulations} \vspace{-10pt}
A key challenge in analyzing black hole EHT images is the highly variable environment of the accretion flow, and the plasma around a black hole is analytically intractable. GRMHD simulations therefore serve as the current gold standard for self-consistently modeling black hole accretion flows~\citep{Porth_2019}. The resulting 3D plasma states are rendered into 2D image-plane intensities by general-relativistic ray tracing, through integration of the radiative transfer equations~\cite{ipol_paper}. Two key physical parameters that characterize the morphology and dynamics of the flow in the image plane are: first, the viewing inclination angle $i$, which sets the angle between the observer's line of sight and the angular momentum vector of the accretion disk; and second, the dimensionless black hole spin $a_*$, which affects both the emission morphology and the photon trajectories near the event horizon that shape the observed image. The movies have a time resolution of 5\M  and an angular resolution of 0.5 $\mu$as. More information is available in \cref{app:background} ~\cite{dhruv_wong_2025, wong_patoka_2022, sgrA_paper_v}.


\paragraph{Dynamical Systems Attractor and Dissipativity} \vspace{-10pt}
Many complex physical processes, including turbulent flows, can be described as dissipative dynamical systems ~\cite{milnor1985concept}. A key property of such systems is that their long-term behavior collapses from a high-dimensional state space onto a lower-dimensional, bounded set known as a global attractor. This attractor represents the stable, invariant manifold containing all physically-plausible states of the system's dynamics ~\cite{li2021learning}. Crucially, these systems possess a \textit{basin of attraction}: trajectories originating from a larger set of initial conditions, even \textit{off-manifold} states such as a blurry image, are rapidly "pulled" onto the attractor by the system's evolution ~\cite{eckmann1985ergodic, dutta2023attractor}. This dissipative "pull" is the theoretical basis for our model's super-resolution. Furthermore, once a trajectory is on this stable manifold, it remains there, which is what enables stable, long-horizon forecasting.

\section{The \codename Framework}
\label{sec:method}
In this section, we introduce \codename as a framework that first forecasts dynamics, then extracts plasma features, and finally infers black hole parameters. 

\subsection{GRMHD Dynamics Forecast Model}
The core of \codename is a dynamics forecast model whose goal is to predict the next state (or frame) of the black hole accretion flow from the current state. This is equivalent to learning a mapping $f(x_t) \mapsto x_{t+1}$, where $x_t \in X$ represents all possible states of the flow. While many neural architectures for spatio-temporal forecasting exist, including powerful models such as transformers and neural operators, we select the U-Net architecture for its data-efficiency and inherent convolutional inductive bias for capturing local turbulent fluctuations in GRMHDs ~\cite{ronneberger2015u, li2020fourier, vaswani2017attention}. 

A key requirement of the dynamics model is to perform stable, autoregressive prediction for $N$ iterations $(f^N(x_0))$ from a blurry initial state $x_0$. The model should simultaneously super-resolve this initial state and maintain stable, physically-plausible dynamics over a long time horizon. In dynamical systems terms, long-term predictions are expected to approach and remain within the global attractor $A$~ \cite{li2021learning}. However, a vanilla U-Net trained on the default $\ell_2$ loss rapidly accumulates error and diverges. As a preliminary step, we log-normalize the states to compress the dynamic range for numerical stability. More crucially, we introduce a multi-scale loss structure for training.

\paragraph{Multi-Scale Laplacian Pyramid Loss} \vspace{-10pt}
Standard pixel-wise losses such as $\ell_2$ treat all spatial frequencies equally, which makes them poorly suited for preserving the holistic disk structure and fine-scale spiral features present in GRMHD frames. To encourage consistency across frequencies, we adopt a Laplacian pyramid loss that applies supervision at multiple spatial scales.  Such multi-scale objectives have proven effective in stabilizing training and preserving structure in vision tasks~\cite{ghiasi2016laplacian}.

Given $M$ levels and downsample (average pooling), bilinear upsample operations, for level $k \in \{0, \ldots M\},$
\begin{equation}
\begin{split}
    Lap_k(I) = G_k(I) - \text{Upscale}(G_{k+1}(I)) 
    \\
    G_k(I) = \text{Downscale}(G_{k-1}(I)), G_0(I) = I
\end{split}
\label{eq:laplacian}
\end{equation}
A standard Laplacian pyramid on simulation frames has 7 layers, i.e. $k\in\{0, \ldots 6\}$. Here, we make two key modifications: first, to save computation, we use a subset of the pyramid levels $k \in \{0,1,2,6\}$ to include both the finest details $(k=0)$ and intermediate-scale information $(k=1,2)$. However, computing the coarsest, single-pixel level $Lap_6 = G_6$ requires the previous level of the Gaussian pyramid, $G_5$. We make the second modification by replacing the coarsest level information $Lap_6$ with the mean flux, $\Phi(I)$, such that it can be computed independent of other levels. For an image with dimensions $H, W$,
\begin{equation}
    Lap_6(I) \coloneq  \Phi(I) = \frac{1}{HW} \sum_{i=1}^H \sum_{j=1}^W I(i,j)
    \label{eq:mean_flux}
\end{equation}

The mean flux, $\Phi(I)$, represents the spatially-integrated intensity of a single frame. A time-series of mean flux values construct the light curve, a crucial observable in many astrophysics topics beyond black holes, such as exoplanets and supernovae \cite{grassberg1971theory, winn2008transit, schnittman2006light}. For black holes, the mean flux's relationship to the light curve makes it an important observable that constrains the entire energy output of the system. Thus, it is a physical constraint that complements the spatial structure encoded in other levels. 

In summary, we construct the Laplacian pyramid for both prediction and ground truth to compute the loss term as a weighted sum of losses for each level:
\begin{equation}
    \mathcal{L}_{total} = \sum\mathcal{L}_{Lap_0} + \frac{1}{2}\mathcal{L}_{Lap_1} + \frac{1}{4}\mathcal{L}_{Lap_2} + \frac{1}{8}\mathcal{L}_{\Phi}
    \label{eq:loss}
\end{equation}
The weights follow a heuristic scale factor and were not hand-tuned~\cite{burt1987laplacian}. 
At each scale, the loss term combines $\ell_1$ and $\ell_2$ with equal weight to balance the outlier robustness of $\ell_1$ and ease of optimization of $\ell_2$. 

\begin{figure}
    \centering
    \includegraphics[width=0.95\linewidth]{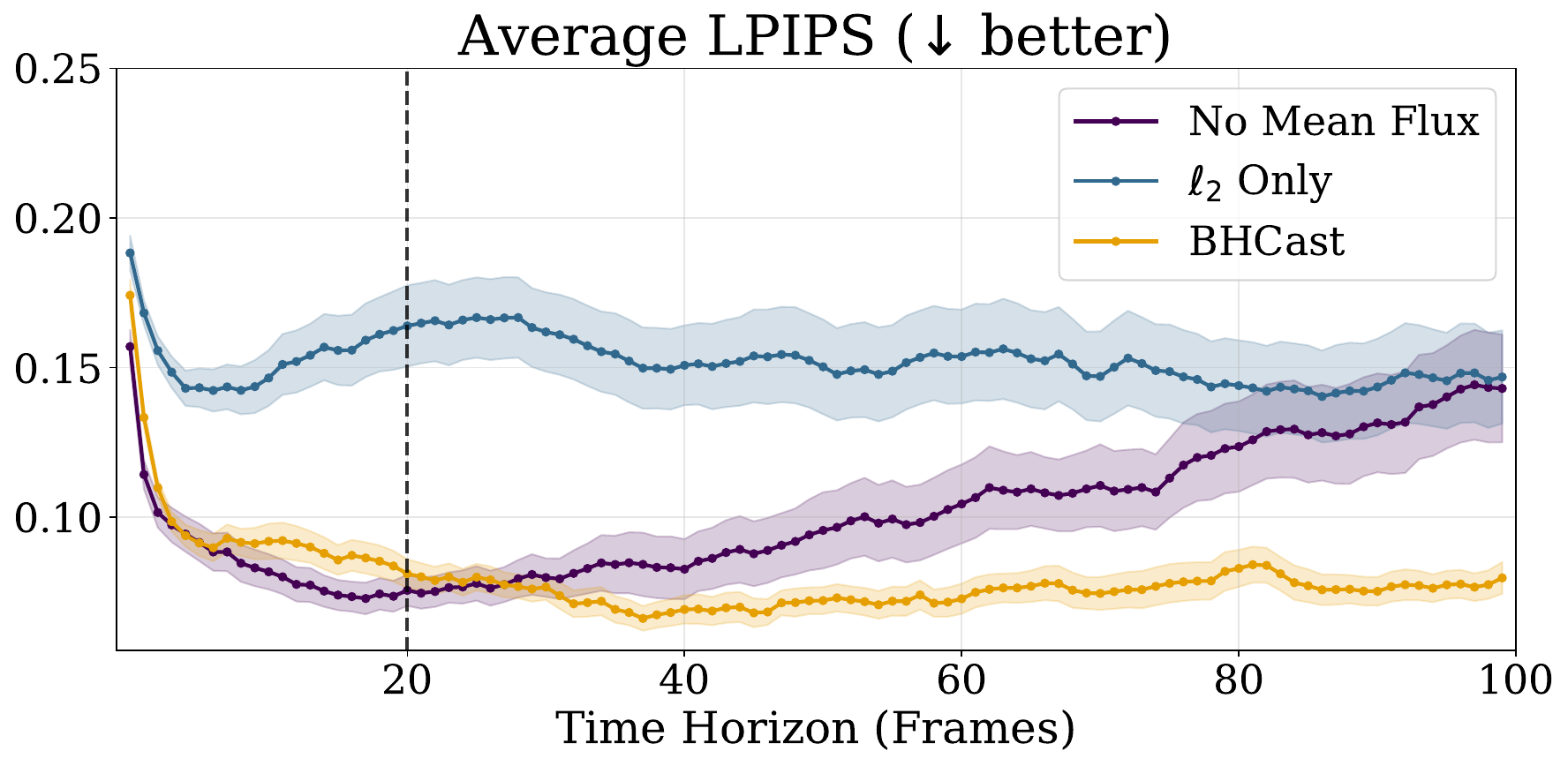}
    \caption{\textbf{Loss ablation study.} LPIPS across three losses: $\ell_2$ only, multi-scale without mean-flux, and the full \codename loss. $\ell_2$-only performs poorly at all times, while the multi-scale variant works briefly but becomes unstable. The full loss, combining multi-scale and mean-flux terms, yields stable long-term forecasts.}
    \label{fig:loss_ablation}
\end{figure}

To evaluate the fidelity of the restored details, we use the Learned Perceptual Image Patch Similarity (LPIPS, lower is better) metric, ensuring that recovered fine-grained structure reflects meaningful physical variability rather than noise \cite{zhang2018unreasonable}. LPIPS measures perceptual similarity beyond pixel-wise metrics like PSNR or SSIM. We perform an ablation study on the dynamics model's training loss. We train two additional dynamics models: (i) using only $\ell_2$ and (ii) only Laplacian pyramid without the mean flux component. \cref{fig:loss_ablation} shows these variants fail in long-term stable forecasting. The multi-scale, no-mean-flux model is only stable for the first 20 forecasts. This confirms the importance of mean flux regularization for long-term forecasting.

\subsection{Extracting Plasma Features from Forecast}
Following the autoregressive generation, \codename extracts plasma features from the forecasted frames using \textbf{cylinder plots}, which capture the dynamical and geometric properties of the accretion flow~\cite{conroy2023rotation}. A cylinder plot $T(\theta, t)$ is a 2D function parameterized by position angle $\theta$ and time $t$, computed by sampling image intensities along a ring. The ring radius $\sqrt{27}~GMc^{-2}D^{-1}$ \footnote{D denotes distance to the black hole} is chosen as it approximately corresponds to the critical impact parameter for Sgr A*, which defines the visible edge of the black hole shadow. Beyond the primary radius, we evaluate additional cylinder plots at multiple radii,  enabling extraction of four quantitative features: two dynamical, (i) pattern speed and (ii) rotation curve slope, and two geometric, (iii) pitch angle and (iv) asymmetry~\cite{conroy2023rotation, Ricarte_2022_pitch_angle, conroy2025event, asymmetry2026}. The following paragraphs introduce and define each of these features.

\paragraph{(i) Dynamics: Pattern Speed $\Omega_p$} \vspace{-12pt}
The pattern speed, a measure of the apparent angular rotation rate defined through ratios of azimuthal image moments, serves as a statistical descriptor of how the overall structure rotates. Visually, pattern speed is the slope of the brightest feature on a normalized cylinder plot. $\Omega_p$ depends strongly on inclination angle, a key property of interest ~\cite{conroy2023rotation}.

\paragraph{(ii) Dynamics: Rotation Curve Slope} \vspace{-12pt}
Recently, Conroy et al. found that rotation rates measured at different distances from the center of the dominant Sgr A* ring exhibit a \textit{rotation curve}. Rotation speeds are faster closer to the black hole event horizon. The rate of change of the rotation speed at different angular radii, the \textit{rotation curve slope}, varies with the black hole's spin parameter \cite{conroy2025event}.

\paragraph{(iii) Geometry: Pitch Angle $\Phi$} \vspace{-12pt}
This quantifies the geometry of coherent spiral structures in the flow by measuring how tightly wound the emission arms are around the black hole shadow ~\cite{Ricarte_2022_pitch_angle}. It is estimated by comparing the azimuthal structure at neighboring radii in the cylinder plot and identifying the relative angular shift between them. Computationally, this is done by cross correlating two cylinder plots measured at $r_{ring}$ and $r_{ring} + \delta_r$ ~\cite{baubock_spiral_prep}. Pitch angle is correlated with the inclination angle \cite{Ricarte_2022_pitch_angle, conroy2023rotation}. 

\paragraph{(iv) Geometry: Asymmetry} \vspace{-12pt}
Asymmetry is the amplitude of the sinusoidal function fitted to the time-averaged cylinder plot. It measures the difference in brightness between the bright side of the ring and the dim side of the ring. This observable feature correlates with spin of the black hole ~\cite{asymmetry2026}. 

\subsection{Physics Inference Model}
The four features from the previous analysis create an information bottleneck, where plasma features identified as important by astrophysicists are distilled from a forecasted simulation. This physics-informed inductive bias enables more data-efficient and robust parameter estimation. The final stage of \codename's pipeline is a classification model that takes a set of plasma features as input, and outputs a category for the spin parameter $a_*$ and the inclination parameter $i$. To train the model, we generate a dataset with plasma features computed on slices of the GRMHD training movies combined with features extracted from forecasts. 

Instead of a deep neural network, we choose a popular gradient-boosting decision tree framework, XGBoost, to perform the feature classification task. The key consideration here is the quantitative feature importance scores and predictive confidence scores from XGBoost, which offer interpretable parameter inference results~\cite{chen2016xgboost}. Our experiments identify pattern speed as most important for inclination classification and asymmetry for spin classification. Full importance score metrics are listed in \cref{app:results}. In addition to interpretability, gradient boosting decision trees are well suited for handling heterogeneous tabular data and are invariant to the scale of different features \cite{mcelfresh2023neural}.

\begin{figure*}
    \centering
    \includegraphics[width=\linewidth]{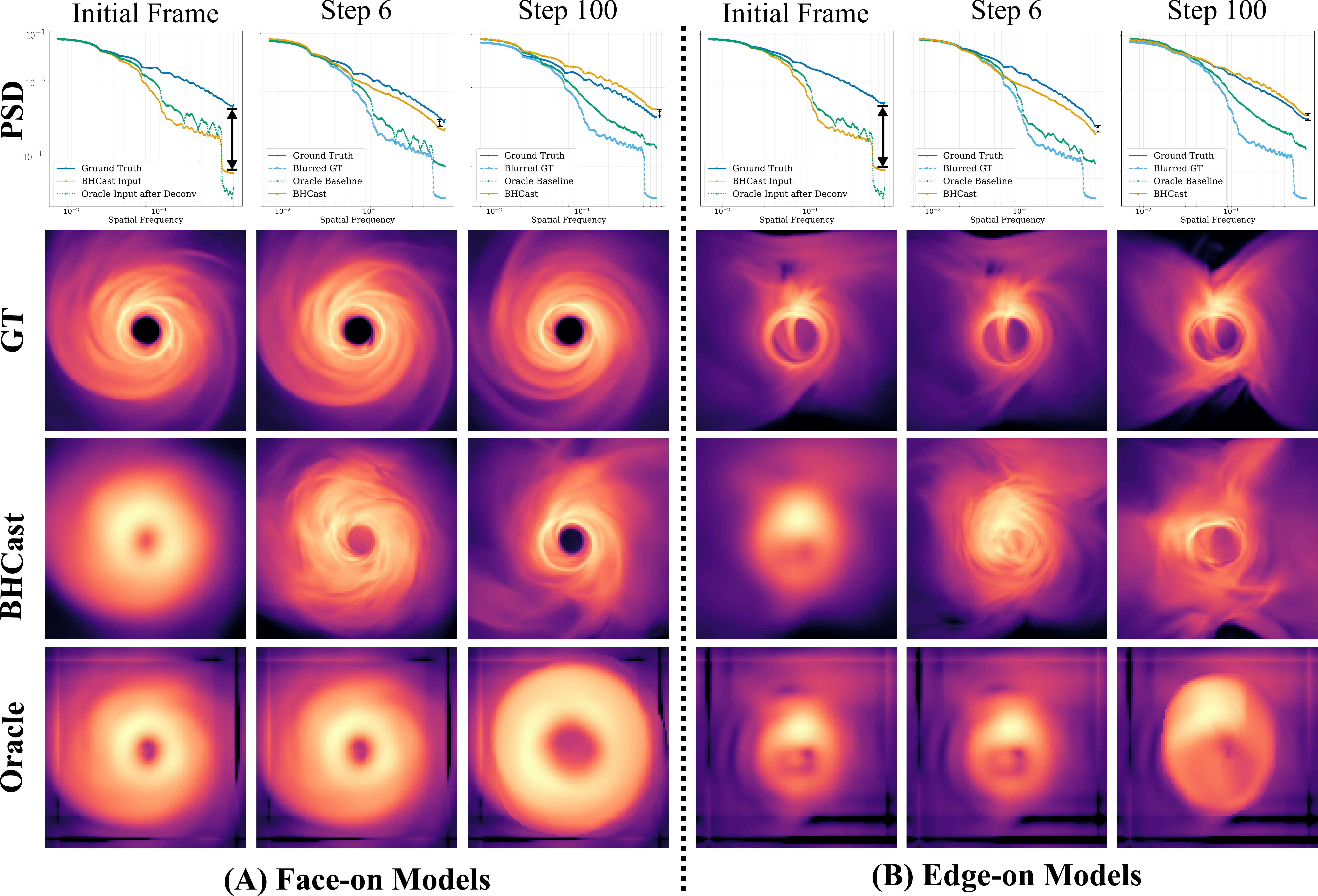}
    \caption{\textbf{Super-Resolution Evaluation}: Power Spectral Density (PSD) and log-scale visualizations of the input, prediction after 6 steps (30 \M), and prediction after 100 steps (500 \M) for (A) face-on GRMHD and (B) edge-on GRMHD. At input, \codename lacks high-frequency details but recovers them by step 6 to match the ground truth PSD. Its forecasts also visually converge to the ground truth. The oracle optical flow baseline starts with the deconvolution result but shows limited super-resolution. The learning-based baseline is omitted due to divergence after a few steps.} 
    \label{fig:forecast_psd}
\end{figure*}

\section{Forecasting Sgr A*}
\label{sec:sgra_results}
We empirically study our proposed framework on the Sgr A* dataset through a three-part evaluation:
\begin{enumerate}
    \item Most importantly, we demonstrate that our dynamics surrogate model super-resolves a blurry input image and performs long-term stable forecast. 
    \item From the forecast, we extract plasma features and show that they are consistent with the ground truth. 
    \item Finally, we infer black hole parameters and demonstrate the robustness of \codename to the input image resolution. 
\end{enumerate}
Note that \codename accomplishes all these tasks within a single inference pass, whereas each baseline handles a single task and must be trained independently.

\subsection{Super-Resolution and Long-Term Forecasting}
This subsection focuses on results for super-resolving a blurry input and long-term autoregressive forecasting. 
\paragraph{Baseline}
\vspace{-10pt}
We compare to two alternative forecasting approaches, one deep-learning and one non-learning baseline. While our model simultaneously super-resolves and evolves dynamics, prior research has developed tools that specialize in either super-resolution or forecasting. Thus, our baselines decouple the two tasks. The \textbf{learning baseline} first uses a pretrained enhanced deep super-resolution network (EDSR), finetuned on GRMHD data, to super-resolve the blurry input image; then, a convolutional LSTM (ConvLSTM) trained on unblurred simulations uses the output from EDSR to forecast dynamics \cite{lim2017enhanced, shi2015convolutional,torchsr}. The non-learning baseline consists of a Wiener deconvolution module followed by an optical flow model to evolve the image \cite{wiener1964extrapolation, farneback2003two}. Given a test frame, we allow access to the ground truth mean optical flow for forecasting. Thus, the non-learning pipeline is an \textbf{oracle baseline}.

\paragraph{Super-Resolution} \vspace{-12pt}
Conventional super-resolution methods operate in a one-shot manner: given a blurry image, the model directly predicts a sharper version in a single step. In contrast, our dynamics-based approach sharpens the image through time: the first predicted frame is noticeably clearer than the input, and each subsequent autoregressive forecast becomes progressively sharper. This behavior arises because the GRMHD simulations form a global attractor -- a manifold of physically-plausible states (frames) in the training data. A blurry input, lacking high-frequency detail, lies off this manifold. The trained dynamics model projects the input back onto the manifold, effectively \textit{pulling} forecasts into the space of physically realistic states and restoring high-frequency detail. Subsequent predictions remain stable and physically consistent. To evaluate super-resolution, we adopt the power spectral density (PSD) as a measure of spatial detail~\cite{robinson2006statistical}. A good super-resolved forecast should match the ground-truth PSD. As shown in \cref{fig:forecast_psd}, averaged across all test cases, \codename begins from a low-frequency, blurry input, unlike baselines that super-resolve before forecasting, yet after six forecast steps its PSD recovers the missing high-frequency content. \cref{fig:forecast_psd} illustrates the PSD result with two forecast test cases, one for face-on GRMHD models (top-down view) and one for edge-on (side view). In both cases, \codename super-resolves the dominant ring by step 6 and maintains spiral arm features that resemble the ground truth until the end of the forecast.

\begin{figure}[t]
    \centering
    \includegraphics[width=\linewidth]{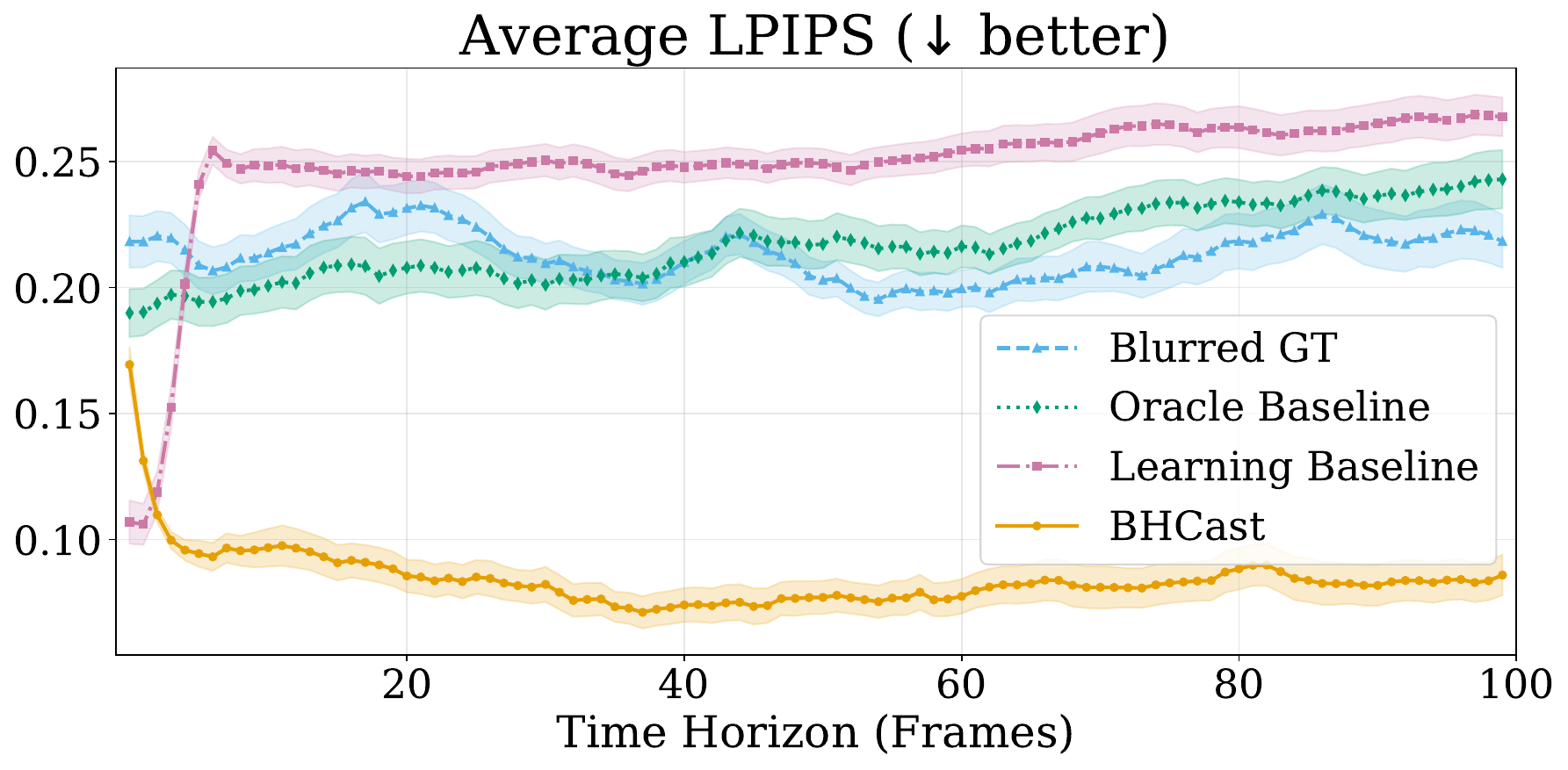} \\
    \caption{\textbf{Forecasting Fidelity Evaluation.} Perceptual similarity (LPIPS; lower is better) between forecasts and the unblurred ground-truth simulation. \codename's forecast remains visually close to the ground truth over long horizons, while baseline forecasts are on par with the EHT-resolution blurred input.}
    \label{fig:forecast_lpips}
\end{figure}

\paragraph{Forecast Fidelity} \vspace{-10pt}
 \cref{fig:forecast_lpips} shows the LPIPS over 100 forecast steps, averaged over test inputs. The learning baseline deteriorates rapidly, while \codename achieves the lowest LPIPS and preserves dynamics throughout the roll-out.
 
\subsection{Plasma Feature Analysis}
This subsection evaluates the accuracy of plasma features extracted from \codename's forecast. 
\paragraph{Baseline}\vspace{-15pt}
As a point of comparison, we consider a supervised regression model that estimates the features directly from the input image, bypassing the cylinder-plots analysis. We train a ResNet50 model similar in parameter count to our U-Net surrogate on (image, feature) pairs, where features are computed from the training set \cite{he2016deep}. Since we compute four distinct features, we treat them as individual tasks and train four separate ResNet models, while \codename extracts all features from a single forecasted movie. 

As a powerful feature extractor, the ResNet baseline achieves almost perfect mean absolute error (MAE) on the validation set for each regression task. However, when tested on blurry images, the distribution shift leads to increased errors. \cref{tab:feature_mae} compares the MAE with standard error (SE) of \codename to the ResNet baselines. Despite accumulating error inherent to autoregressive prediction, \codename extracts pattern speed $\Omega_p$ significantly better, while remaining on par with direct estimation for other features. 

\begin{figure}[h]
    \centering
    \includegraphics[width=\linewidth]{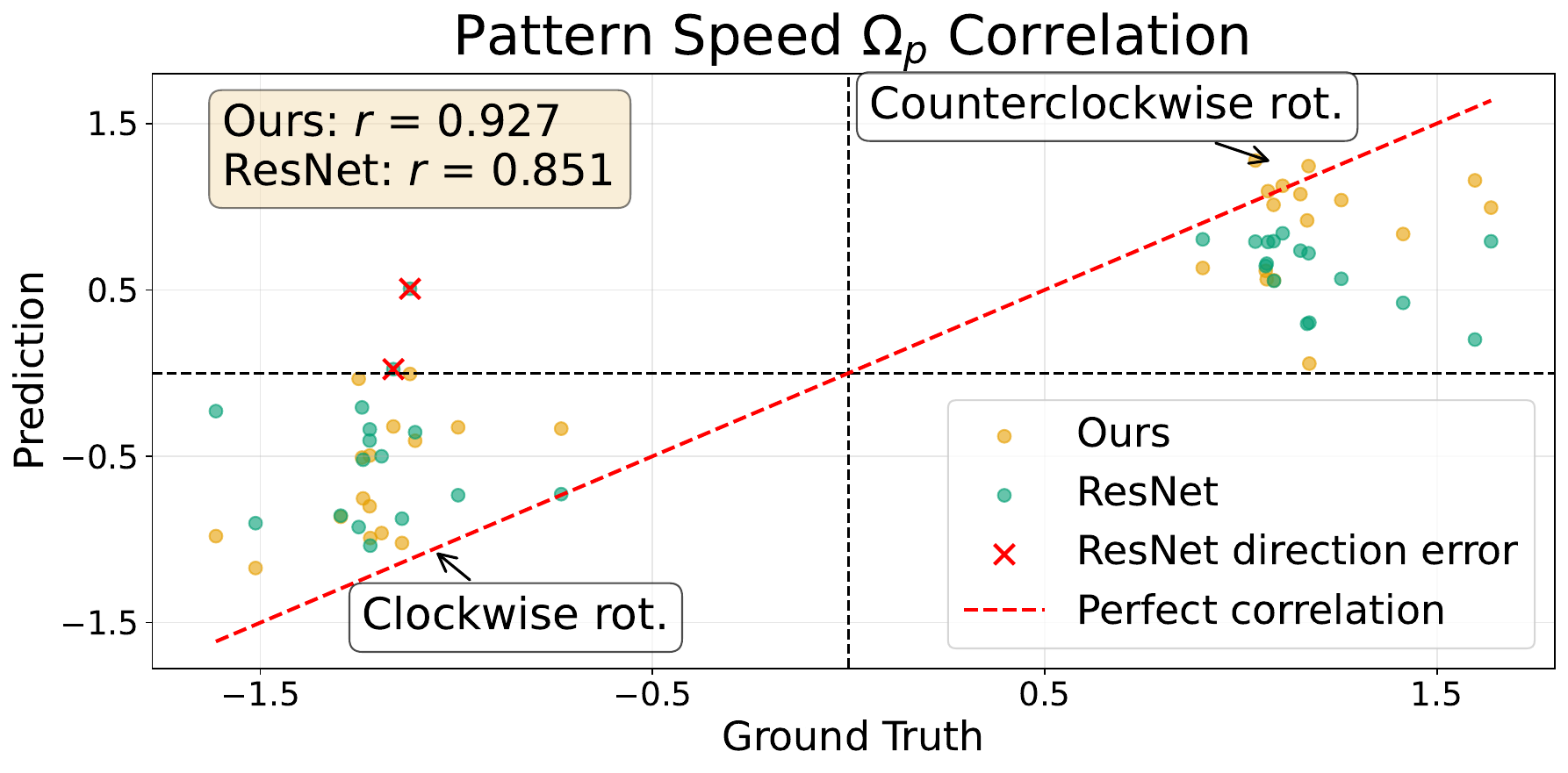} \\
    \caption{\textbf{Pattern Speed Correlation}. Extracted pattern speed from a 60-frame forecast shows a very strong correlation of 0.927 with the ground truth values. In addition, \codename identifies separate clusters for clockwise and counter-clockwise rotation, showing significantly less bias than the baseline when estimating counterclockwise rotation.
    }
    \label{fig:feature_correlation}
\end{figure}

In addition to reporting the absolute error, \cref{fig:feature_correlation} demonstrates strong correlation of $0.927$ between \codename's extracted and ground truth pattern speeds. Our estimates of counter-clockwise rotation speeds are less biased than the baseline. For the remaining plasma features the correlations are $0.758$ for pitch angle, $0.626$ for asymmetry, and $0.610$ for rotation curve slope. Correlation plots for these additional features are provided in \cref{app:results}.  

\begin{table}[t]
\centering
\caption{\textbf{Mean Absolute Error $\pm$ Standard Error ($\downarrow$) of extracted plasma features, averaged across all test cases.} Our BHCast model is compared against a ResNet baseline.}
\label{tab:feature_mae}
\begin{tabular}{lcc}
\toprule
\textbf{Plasma Feature} & \textbf{\codename} & \textbf{ResNet Baseline} \\
\midrule
Pattern Speed ($\Omega_p$) & \textbf{0.46 $\pm$ 0.05} & 0.64 $\pm$ 0.07 \\
Pitch Angle ($\Phi$)      & \textbf{0.13 $\pm$ 0.01} & 0.14 $\pm$ 0.02 \\
Asymmetry                & 0.30 $\pm$ 0.03         & \textbf{0.23 $\pm$ 0.02} \\
Rotation Curve Slope     & \textbf{0.24 $\pm$ 0.03} & 0.25 $\pm$ 0.03 \\
\bottomrule
\end{tabular}
\end{table}

\subsection{Black Hole Parameter Inference}
This subsection presents the parameter inference accuracies and robustness to shifts in the input image distribution.
\paragraph{Baseline} \vspace{-10pt}
Since our GRMHD data only covers \textbf{discrete values of spin and inclination}, a convolutional neural network can directly predict the class from an image. We train two additional ResNet50 models, one for inclination and one for spin, until convergence. There are two classes for spin prediction: prograde ($a_* >0)$ and retrograde ($a_*<0$), and four classes for inclination prediction: negative face-on, negative edge-on, positive edge-on, positive face-on. 

At test time, we vary the blurring kernel to differ from the standard EHT-resolution of 20 $\mu as$ to include 25 $\mu as$ and 30 $\mu as$ options for testing robustness to distribution shift. Evaluated on 640 samples, \cref{tab:parameter_accuracy} demonstrates the superior accuracy of \codename with the standard $20 \mu as$ EHT-resolution blurring. Moreover, changing the blurring kernel leads to massive degradation of the supervised baseline. In contrast, \codename is robust to different blurring kernels, due to effective super-resolution of the dynamics model and the accuracy of plasma feature estimation. We include additional robustness studies in \cref{app:results}.

\begin{table}[t]
\centering
\caption{\textbf{Classification accuracy of black hole parameters, comparing \codename to a ResNet baseline.} The 20 $\mu$as (EHT-like) resolution is the baseline. The 25 $\mu$as and 30 $\mu$as rows show the performance degradation (or stability) as blurring increases, with differences ($\Delta$) relative to the baseline.}
\label{tab:parameter_accuracy}
\resizebox{\linewidth}{!}{ 
\begin{tabular}{llcc}
\toprule
Blurring Level & Model & Inclination Accuracy & Spin Accuracy \\
\midrule
\multirow{2}{*}{\shortstack{20 $\mu$as \\ (Baseline)}} & \textbf{\codename} & \textbf{56.41\%} & \textbf{69.22\%} \\
                            & \textbf{ResNet}        & 47.19\% & 67.66\% \\
\midrule
\multirow{2}{*}{25 $\mu$as} & \textbf{\codename} & \textbf{56.72\% (\textcolor{green}{+0.31})} & \textbf{71.09\% (\textcolor{green}{+1.87})} \\
                            & \textbf{ResNet}        & 31.41\% (\textcolor{red}{-15.78}) & 54.53\% (\textcolor{red}{-13.13}) \\
\midrule
\multirow{2}{*}{30 $\mu$as} & \textbf{\codename} & \textbf{53.44\% (\textcolor{red}{-2.97})} & \textbf{65.78\% (\textcolor{red}{-3.44})} \\
                            & \textbf{ResNet}        & 25.47\% (\textcolor{red}{-21.72}) & 44.37\% (\textcolor{red}{-23.29}) \\
\bottomrule
\end{tabular}
} 
\end{table}

\paragraph{Uncertainty Quantification via Bootstrapping} \vspace{-10pt}
We perform bootstrapping by training an ensemble of 100 models on random $80\%$ portions of the training set of plasma features. On the held-out validation set, we measure the ensemble's \textit{aleatoric uncertainty} via average score entropy and \textit{epistemic uncertainty} via average variance ~\cite{houlsby2011bayesian, hofer2002approximate}. In \cref{tab:uncertainty}, the extremely-low epistemic uncertainties for both spin and inclination prediction indicate that the ensemble unanimously agrees on predictions. The errors stem from a small portion of data outliers as aleatoric uncertainty. 

\begin{table}[b]
\centering
\caption{\textbf{An ensemble of XGBoost bootstrapped models achieves high accuracy with low epistemic uncertainty.} This demonstrates that the extracted plasma features are effective and lead to stable, unanimous predictions.}
\label{tab:uncertainty}
\resizebox{\linewidth}{!}{

\begin{tabular}{l c c}
\toprule
\textbf{Metric} & \textbf{Spin Model} & \textbf{Inclination Model} \\
\midrule
Validation Accuracy & 0.94 & 0.87 \\
Aleatoric Uncertainty & $0.15 (\in [0, 1]) $ & $0.42 (\in [0,2])$ \\
Epistemic Uncertainty & $\textbf{0.003} (\in [0,1])$ & $ \textbf{0.004} (\in [0,1]$) \\
Mean Confidence & $0.96 \pm 0.10$ & $0.88 \pm 0.15$ \\
\bottomrule
\end{tabular}
}
\end{table}

\paragraph{Dataset and Experiment Details} \vspace{-10pt}
Our GRMHD dataset consists 32 Sgr A* movies, each with a temporal window of 1,000 simulation frames with resolution $100 \times 100$ pixels. We partition the movies into 800/100/100 frames for training/validation/test sets respectively. For dynamics forecasting, a U-Net is trained on consecutive GRMHD frames from the training set, totaling 25,000 frame pair samples. For parameter inference, an XGBoost model is trained on plasma features computed on consecutive 60-frame slices from the training set and 60-frame forecasts from the trained U-Net, and the source GRMHD movie spin and inclination are used as labels. Over 30,000 samples of (plasma features, black hole label) data are generated for training. During testing, input frames are blurred to EHT resolution and two frames apart (10 \M) within the test set. The time horizon for test-time forecast is 60 frames (300 \M), such that we balance having sufficient evaluation samples and a significant length for accurate plasma features. For hyperparameters and model specifications, refer to \cref{app:pipeline}.

\section{Forecasting M87*} \vspace{-5pt}
\label{sec:m87_results}
In this section, we test the generalization of \codename by applying it to a new black hole system, M87*. Since M87* GRMHD data is very limited compared to Sgr A*, it is challenging to train a standalone dynamics model for M87*. 
\vspace{-5pt}

\subsection{M87* Parameter Estimation} \vspace{-5pt}
Previous VLBI measurements have estimated the inclination angle of M87* as either 17\textdegree or 163\textdegree ~\cite{walker2018structure}. The ambiguity between the two options remains due to uncertainty as to whether the accretion disk angular momentum is pointed towards or away from Earth. Given a blurry test M87* GRMHD image, we leverage an important heuristic that pattern speed has the same sign as $\cos(\text{inclination})$ to constrain this value ~\cite{conroy2023rotation}. We produce a forecast from the input image and compute the pattern speed to check its sign. We classify input images with positive predicted pattern speeds to 17\textdegree and negatives to 163\textdegree. On 168 forecasts, \codename achieves \textbf{78.6\%} accuracy on negative samples and \textbf{58.3\%} on positive samples without any training on M87* data.  

\subsection{Testing on Real M87* Images} 
\begin{figure}
    \centering
    \includegraphics[width=\linewidth]{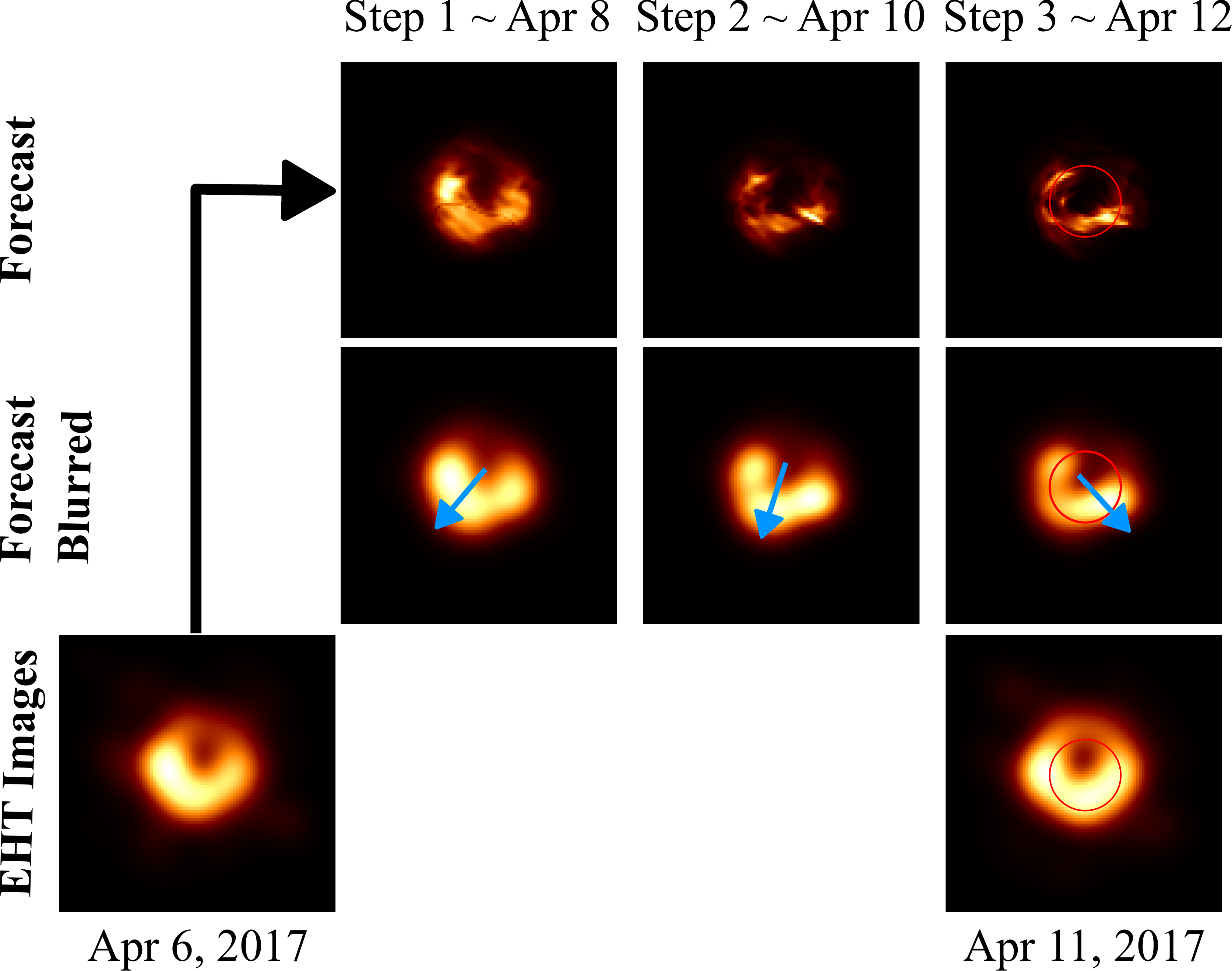}
    \caption{\textbf{Extrapolation on EHT M87* image}. We use the April 6 EHT image as input and validate our forecasting ability by comparing the blurred forecast after three autoregressive steps with the April 11 EHT image. \codename successfully resolves the correct ring structure (red circle) and predicts a counterclockwise rotation, consistent with the EHT reconstructions.}
    \label{fig:m87_extrapolation}
\end{figure}

Although \codename was trained only on simulations and not on EHT reconstructions, we evaluate it on real observational data: the EHT images of M87*. The April 2017 EHT campaign captured M87* over multiple days; the April 6 and April 11 reconstructions are separated by approximately 15 \M, corresponding to 3 steps of our dynamics forecast model. \cref{fig:m87_extrapolation} shows our forecast result resolving a stable ring structure with the correct radius. Moreover, our model predicts a counterclockwise brightness shift along the ring, consistent with the observed April 11 EHT image. Additionally, we achieve 68.9\% accuracy in rotation prediction on the augmented test set of M87* images in \cref{tab:m87_tta}.
 \vspace{-5pt}

\section{Conclusion and Discussion}
\label{sec:conclusion}

We introduced \codename, a framework that forecasts black hole dynamics from a single, blurry image. Using a physics-informed multi-scale loss, our dynamics surrogate model produces both super-resolved and stable predictions over long temporal horizons. From these forecasts, we extract physically meaningful plasma features, and use them to infer key properties of the black hole system. \codename therefore enables dynamical analysis even from highly blurred EHT images. Looking ahead, we plan to extend \codename's inference capabilities to the multi-day M87* movie campaign. By recasting a traditionally static inverse problem into a modular task of stable forecasting and physical inference, \codename suggests a new paradigm for scientific imaging, where dynamics are not just a byproduct of observation, but a foundational tool for discovery.

\clearpage 

\section*{Acknowledgment}
 We thank Rohan Dahale for assistance with plasma feature computation and the PI-vision group/EcoSys group at University of Toronto for useful feedback on the paper. We thank Kamyar Azizzadenesheli for his important role in guiding the project. We thank Charles Gammie from UIUC for providing crucial data used in this work. 
 
 \noindent RT and GP are supported by the Canada Foundation for Innovation JELF grant, NSERC Discovery Grant, Google Scholar Research Award, VMware Early Career Faculty Grant/Award, Ontario Early Researcher Award, and the Vector CIFAR AI Chair award. NC is supported by the NASA Future Investigators in NASA Earth and Space Science and Technology (FINESST) program. This material is based upon work supported by the National Aeronautics and Space Administration under Grant No. 80NSSC24K1475 issued through the Science Mission Directorate. AS and AL are supported by the Natural Sciences and Engineering Research Council of Canada (NSERC). This work was supported by the Ontario Research Fund – Research Excellence under Project Number RE012-045.
 
 \noindent This work was further supported by NSF grants AST 17-16327 (horizon), OISE 17-43747, and AST 20-34306.  This research used resources of the Oak Ridge Leadership Computing Facility at the Oak Ridge National Laboratory, which is supported by the Office of Science of the U.S. Department of Energy under Contract No. DE-AC05-00OR22725.  This research used resources of the Argonne Leadership Computing Facility, which is a DOE Office of Science User Facility supported under Contract DE-AC02-06CH11357.  This research was done using services provided by the OSG Consortium, which is supported by the National Science Foundation awards \#2030508 and \#1836650.  This research is part of the Delta research computing project, which is supported by the National Science Foundation (award OCI 2005572), and the State of Illinois. Delta is a joint effort of the University of Illinois at Urbana-Champaign and its National Center for Supercomputing Applications.

{
    \small
    \bibliography{cvpr2026_conference}
    \bibliographystyle{ieeenat_fullname}
}

\clearpage

\appendix

\section{\codename Pipeline Details}
\label{app:pipeline}
This section provides model information and hyperparameters used during training and evaluation for each stage of the framework.
\subsection{Dynamics Forecast Model}
First, we report the model details of the U-Net surrogate and other baselines. 

\paragraph{U-Net} \vspace{-8pt}
 The U-Net used in our experiments has the same architectural hyperparameters as those defined in the original paper ~\cite{ronneberger2015u}, based on a standard encoder-decoder design with skip connections. The network comprises four downsampling steps and four corresponding upsampling steps. \\

 \textit{Encoder:}  
 \begin{itemize} 
 	\item DoubleConv blocks: two $3 \times 3$ convolution layers, each followed by batch normalization and ReLU activation. 
 	\item Downsampling: $2 \times 2$ max pooling with stride 2. 
 	\item Number of channels (features) starts from 64 and reaches 1024 at the end of the encoder. \\
 \end{itemize}

\textit{Decoder:}
\begin{itemize}
	\item DoubleConv blocks same as above.
	\item Transposed Convolution: reduces channel size by a factor of 2 with $2 \times 2$ convolution with stride 2. 
	\item Skip connections: outputs from transposed convolution are concatenated with encoder feature maps via skip connections. \\
\end{itemize}

\textit{Output Head:} $1 \times 1$ convolution to map the 64-channel feature map to the number of output channels, which is 1. \\

\noindent All hyperparameters for training are as follows:
\begin{verbatim}
batch_size: 128
optimizer: AdamW
learning_rate: 0.001
weight_decay: 0.0001
scheduler: cosine
t_max: 500
epochs: 500
loss: Multi-scale Laplacian Pyramid
\end{verbatim}

\paragraph{Oracle Baseline} \vspace{-5pt}
Our non-learning baseline consists of a Wiener deconvolution module with an estimated PSF and an optical flow module ~\cite{farneback2003two, wiener1964extrapolation}. Given a test frame, we give the baseline access to the training frames for the same movie to compute the following: PSF estimation starts with a kernel size very close to the ground truth blurring kernel size. Noise-to-Signal Ratio (NSR) for Wiener deconvolution is computed on the training set. Optical flow is the mean of the training set. 

 We list the details below: 
\begin{itemize} \item 
	\textit{PSF Estimation (Calibration):} The Point Spread Function (PSF) is modeled as a Gaussian kernel. The optimal standard deviation is estimated via a grid search that minimizes the Mean Squared Error (MSE) between the convolved unblurred training frames and the observed blurred frames. 
	\item \textit{Wiener Deconvolution:} The initial blurred test frame is restored using Wiener deconvolution in the frequency domain. NSR is estimated by calculating the ratio of the residual noise power to the signal power of the unblurred training frames. 
	\item \textit{Mean Optical Flow:} Optical flow maps are computed between consecutive unblurred training frames using the Farneback algorithm. These fields are averaged over the training set to produce a single, static "Mean Flow" field representing the global dynamics. 
\end{itemize}

Hyperparameters of the oracle baseline: 
\begin{verbatim}
	PSF estimation:
	 kernel_size: 21 
	 sigma_grid: 0.5 to 6.0 (23 steps) 
	 num_samples: 50
	Wiener deconvolution:
	 nsr_clip_min:     1e-6
	 nsr_clip_max:     1e3
	Farneback optical flow:
	 pyr_scale:        0.5
	 levels:           3
	 winsize:          25
	 iterations:       3
	 poly_n:           7
	 poly_sigma:       1.5
\end{verbatim}

\paragraph{Learning Baseline}  \vspace{-10pt}
Our learning baseline consists of a EDSR network for super-resolving the blurry input ($x_t$) and a ConvLSTM network for forecasting dynamics by autoregressively producing N frames $(x_{t+1}, \ldots x_{t+N})$ based on the super-resolved image ~\cite{shi2015convolutional, lim2017enhanced}. The EDSR model is fine-tuned from a pre-trained model, while the ConvLSTM model is trained from scratch. 

Hyperparameters of the learning baseline: 
\begin{verbatim} 
EDSR:
	EDSR variant: 16 Blocks, 64 Filters
	scale_factor: 2 
	residual_scale: 0.1 
	fine_tune_epochs: 5 

ConvLSTM:
	layers:           2
	hidden_channels:  [64, 64]
	kernel_size:      3x3
	training_epochs:  30
	
training:
	optimizer:        AdamW
	learning_rate:    1e-4
	batch_size:       128
	input_seq_len:    1
\end{verbatim}

\paragraph{Compute Comparison} \vspace{-5pt}
In \cref{tab:forecast_compute}, we compare the parameter count and GFLOPs count between the learning baseline and \codename's dynamics forecast U-Net. Though ConvLSTM has a lower parameter count, the overall FLOPs of the baseline far exceed the U-Net, which does not involve LSTM cell gates. For training, we provide the same compute budget of 5 GPU hours for the EDSR fine-tuning and ConvLSTM training to match that of the U-Net. 

\begin{table}[h]
	\centering
	\caption{Comparison of models in terms of trainable parameters and floating-point operations (FLOPs).}
	\label{tab:forecast_compute}
	\resizebox{\columnwidth}{!}{%
		\begin{tabular}{lcc}
			\toprule
			\textbf{Model} & \textbf{Parameters} & \textbf{FLOPs (G)} \\
			\textbf{EDSR (Super-Resolution)} & $1.37$ M & $6.87$ \\
			\textbf{ConvLSTM (Dynamics)} & $0.45$ M & $17.81$ \\
			\textbf{EDSR + ConvLSTM (Combined)} & $1.82$ M & $24.68$ \\
			\midrule
			\textbf{\codename (Ours - U-Net)} & $31.04$ M & $8.12$ \\
			\bottomrule
		\end{tabular}%
	}
\end{table} 

\paragraph{Additional Modern Baselines: FNO and Diffusion} \vspace{-5pt}
We compare \codename to two additional deep learning baselines representing state-of-the-art approaches in dynamical systems modeling. Different from the aforementioned multi-stage baselines, the Fourier Neural Operator(FNO) and Diffusion models are single-stage and perform end-to-end frame generation like the U-Net \cite{li2020fourier, ho2020denoising}. 

Hyperparameters of FNO:
\begin{verbatim}
n_modes: (128, 128)
hidden_channels: 64
n_layers: 6
in_channels: 1
out_channels: 1
projection_channel_ratio: 2
norm: group_norm
epochs: 500
\end{verbatim}

Hyperparameters of 3D-Conditional Diffusion:
\begin{verbatim}
seq_len: 10
in_channels: 2 (class-conditioned)
out_channels: 1
block_out_channels: (32, 64, 128, 256)
layers_per_block: 2
num_train_timesteps: 1000
beta_schedule: linear
num_inference_steps: 50
learning_rate: 1e-4
weight_decay: 1e-4
batch_size: 1
gradient_accumulation_steps: 4
epochs: 100
\end{verbatim}

\paragraph{Compute Comparison between Additional Baselines} \vspace{-5pt} We report compute in \cref{tab:additional_compute}, including both model size and total forecasting FLOPs count. We also measure training cost in GPU hours on the same hardware. Most importantly, we compare inference throughput in frames per second (FPS) to measure forecasting efficiency. We find that U-Net is \textbf{27.1} $\times$ faster in inference compared to Diffusion, which guarantees fast roll-outs and ensemble-based uncertainty estimates. The efficiency factor of U-Net is crucial for scaled-up parameter inference in the upcoming M87* EHT movie campaigns that increase the number of reconstructed frames and pipeline variants. 

\begin{table}[h]
\centering
\small 
\caption{\textbf{Compute/throughput comparison.} Diffusion uses 20 denoising steps and is evaluated with FP16 AMP due to memory constraints; other models are evaluated in FP32.}

\vspace{-5pt}
\begin{tabularx}{\columnwidth}{Xcccc}
\toprule
\textbf{Model} & \textbf{Param}    & \textbf{Roll-out} & \textbf{Training} & \textbf{Inference} \\
\textbf{Name}  & \textbf{Count(M)} & \textbf{FLOPs (G)}   & \textbf{GPU Hrs}  & \textbf{FPS}       \\
\midrule
FNO            & 409.01            & 39.63          & 19.1              & 110.45             \\ \addlinespace[2pt]
Diffusion      & 28.64             & 75340.88       & 146.5             & 17.91              \\ \addlinespace[2pt]
U-Net         & 31.04             & 486.95         & 5.5               & \textbf{485.20}             \\
\bottomrule
\label{tab:additional_compute}
\end{tabularx}
\end{table}

\subsection{Plasma Feature Extraction Module}
For the extraction module, we provide details to the plasma analysis for each feature. 

\paragraph{Pattern Speed $\Omega_p$} \vspace{-5pt}
Extracting pattern speed requires a cylinder plot computed at the predefined ring radius $\sqrt{27} GMc^{-2}D^{-1}$. For the cylinder plot $T$, we start by converting pixels on the ring to angular positions across time to a 2D matrix with dimensions (prediction temporal length, number of angular positions). $T$ is then normalized to $\tilde{T}$ through taking its log and subtracting its mean ~\cite{conroy2023rotation}. 

$\Omega_p$ is calculated from the second moments of the autocorrelation function $\xi$ of a normalized cylinder plot $\tilde{T}$, namely: 
	\begin{equation}
		\label{eq: pattern_speed}
		\xi(\Delta t, \Delta \theta) = \frac{1}{\sigma^2}\mathcal{F}^{-1}(|\mathcal{F}(\tilde{T})|^2)
	\end{equation}
	where $\sigma^2$ is the variance of $\tilde{T}$ and $\mathcal{F}$ is the Fourier transform. The ratio of second moments of $\xi$ yields $\Omega_p$.

\paragraph{Rotation Curve Slope} \vspace{-5pt}
In addition to the fixed ring radius for Sgr A*  $r_{ring} = \sqrt{27} GMc^{-2}D^{-1}$, we compute 5 additional pattern speeds measured at heuristic-based factors of the radius $\{0.75, 0.9375, 1.125, 1.3125, 1.5 \} \times r_{ring}$ ~\cite{conroy2025event}. The rotation curve slope is the slope of the best first-order fit to the pattern speeds. 

\paragraph{Pitch Angle $\Phi$}\vspace{-5pt}
$\Phi$ is derived from correlating two cylinder plots, taken at radii $r_{ring}$ and $r_{ring} + \delta_r$, where $\delta_r = 0.1$ ~\cite{baubock_spiral_prep}. 

\begin{equation}
    \label{eq: pitch_angle}
    \theta^* = \underset{\theta}{\mathrm{argmax}}(\xi(\tilde{T}(r_{ring}), \tilde{T}(r_{ring} + \delta_r))
\end{equation}

\begin{equation}
    \Phi = \arctan(\frac{\theta^*}{\ln((r_{ring} + \delta_r)/r_{ring})})
\end{equation}

\paragraph{Asymmetry}\vspace{-5pt}
While previous features require 2D cylinder plots, asymmetry uses a 1D, time-averaged cylinder plot to fit a sinusoidal function  ~\cite{asymmetry2026}. The function has the following form:
\begin{equation}
	\label{eq:asym}
	\begin{split}
	A * \cos(\theta + \theta_0) + C 
	\\
	Asymmetry \coloneqq A/C
	\end{split}
\end{equation}

\paragraph{ResNet50 Baseline} 
We provide model details and training hyperparameters of the supervised regression model, ResNet50: 

\begin{verbatim}
	input_channels: 1 
	output_dim: 1 (Scalar Regression)
	
	training:
	optimizer:        AdamW
	learning_rate:    1e-3
	weight_decay:     1e-4
	batch_size:       128
	epochs:           100
	scheduler:        CosineAnnealingLR
	loss_function:    MSELoss

\end{verbatim}

\begin{table}[b]
	\centering
	\caption{Comparison of parameter count and FLOPs count between our framework and ResNet50. Note that baseline is task-specific; extracting all four physical parameters requires training four separate ResNet models.}
	\label{tab:feature_compute}
	\resizebox{\columnwidth}{!}{%
		\begin{tabular}{lcc}
			\toprule
			\textbf{Model} & \textbf{Parameters (M)} & \textbf{FLOPs (G)} \\
			\midrule
			\textbf{ResNet50 (Single Task)} & 23.50 & 0.99 \\
			\textbf{4$\times$ ResNet50 (All Tasks)} & 94.02 & 3.96 \\
			\midrule
			\textbf{BHCAST (Ours)} & 31.04 & 8.12 \\
			\bottomrule
		\end{tabular}%
	}
\end{table}

\paragraph{Compute Comparison} \vspace{-5pt}
In \cref{tab:feature_compute}, we compare the computation requirements of our approach and the baseline. We did not train on a larger supervised learning model since ResNet50 already achieves near-perfect error on the validation set by memorizing the training data.

\begin{figure}[htp!]
    \centering
    \includegraphics[width=\linewidth]{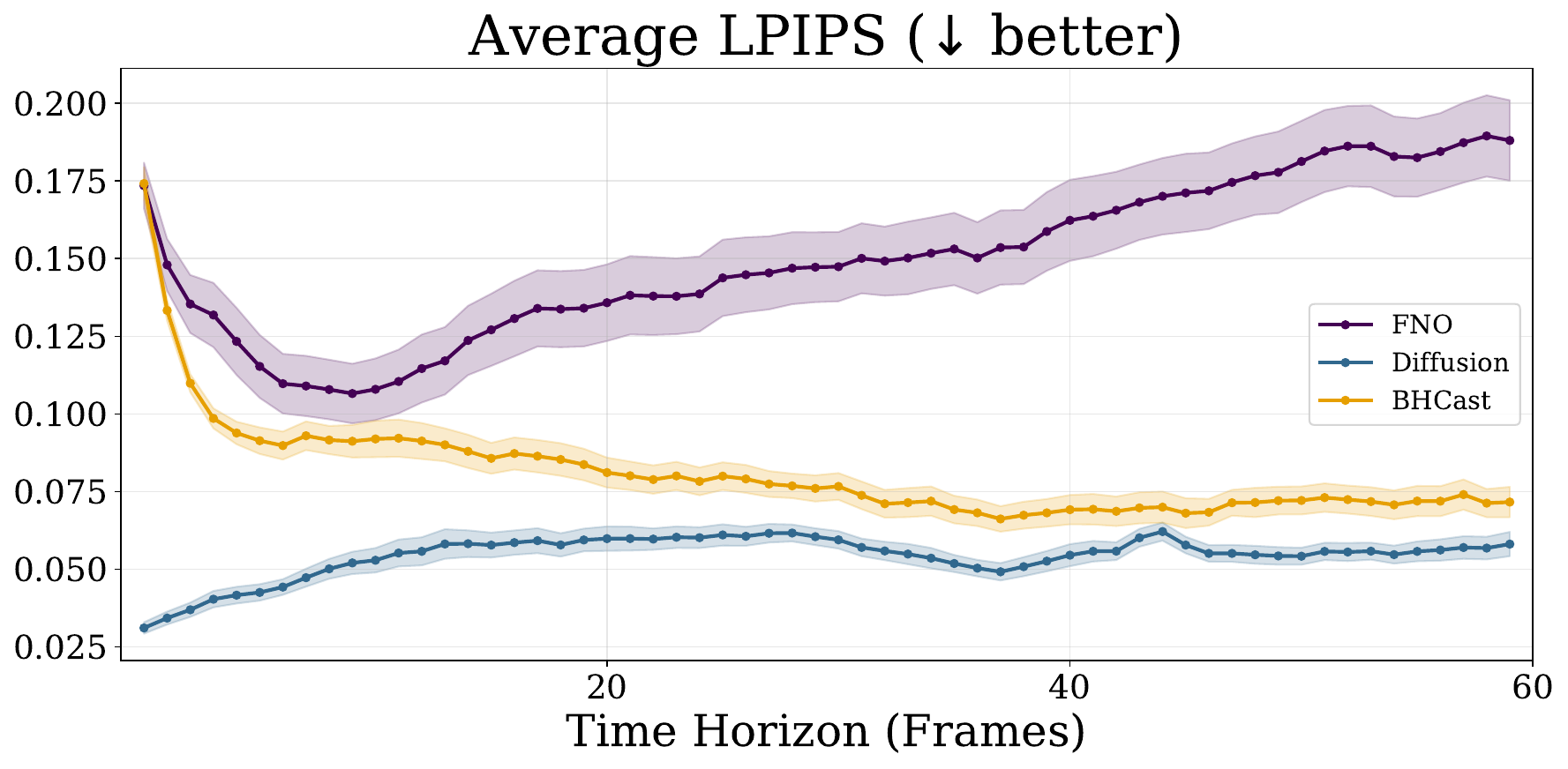}
    \caption{\textbf{Forecasting Fidelity of FNO and Video Diffusion.} }
    \label{fig:lpips_fno_diffusion}
\end{figure}

\subsection{Physics Inference Model}
The hyperparameters used for the XGBoost classifier for physics inference, mostly unchanged from the default values, are as follows: 
\begin{verbatim}
n_estimators:     1000
max_depth:        6
learning_rate:    0.05
subsample:        0.8
colsample_bytree: 0.8
min_child_weight: 1.0
gamma:            0.0
reg_alpha:        0.0
reg_lambda:       1.0
objective:        multi:softprob
eval_metric:      mlogloss
\end{verbatim}

\section{Expanded Empirical Results}
\label{app:results}
To validate \codename, we have conducted extensive empirical experiments, and results from this section should complement those in \cref{sec:sgra_results} and \cref{sec:m87_results}. 

\subsection{Dynamics Forecasting}
For a video visualization of \codename forecasts, refer to our project webpage. We also include a LPIPS comparison between U-Net and models with global inductive biases (FNO/Diffusion) in \cref{fig:lpips_fno_diffusion}, which shows the quality of U-Net forecasts comparable to a compute-intensive diffusion model at long horizons. In addition, FNO diverges in forecast after $\sim$15 steps despite higher model capacity and training with the multi-scale Laplacian loss introduced in \cref{sec:method}.  This suggests that for modeling local turbulent fluxes in GRMHD, the \textbf{local inductive bias} of the U-Net is critical, particularly where data efficiency is required.

\subsection{Plasma Feature Extraction}

\paragraph{Feature Estimation Error} \vspace{5pt}
In \cref{tab:feature_mae_errorbar}, we present the results of feature estimation where \codename is compared to the ResNet50 baseline, as a full version of \cref{tab:feature_mae}.

\begin{table*}
	\centering
	\caption{Estimation MAE of extracted features $\pm$ standard error of extracted plasma features, grouped by different black hole spin $a_*$. This table is the fully expanded \cref{tab:feature_mae}}.
	\label{tab:feature_mae_errorbar}
	\resizebox{\linewidth}{!}{
		\begin{tabular}{lcccccccc}
			\toprule
			& \multicolumn{2}{c}{Pattern Speed $\Omega_p \in \mathbb{R}$} & \multicolumn{2}{c}{Pitch Angle $\Phi \in [0,1]$} & \multicolumn{2}{c}{Asymmetry $\in \mathbb{R}^+$} & \multicolumn{2}{c}{Rotation Curve Slope $\in \mathbb{R}$} \\
			\cmidrule(lr){2-3} \cmidrule(lr){4-5} \cmidrule(lr){6-7} \cmidrule(lr){8-9}
			Spin ($a_*$) & \textbf{\codename} & \textbf{ResNet} & \textbf{\codename} & \textbf{ResNet} & \textbf{\codename} & \textbf{ResNet} & \textbf{\codename} & \textbf{ResNet} \\
			\midrule
			-0.94 & 0.72 $\pm$ 0.11 & 0.80 $\pm$ 0.17 & 0.16 $\pm$ 0.03 & 0.11 $\pm$ 0.04 & 0.23 $\pm$ 0.05 & 0.22 $\pm$ 0.04 & 0.25 $\pm$ 0.05 & 0.29 $\pm$ 0.06 \\
			-0.5  & 0.37 $\pm$ 0.14 & 0.51 $\pm$ 0.08 & 0.08 $\pm$ 0.01 & 0.07 $\pm$ 0.02 & 0.27 $\pm$ 0.04 & 0.15 $\pm$ 0.04 & 0.14 $\pm$ 0.02 & 0.13 $\pm$ 0.01 \\
			0.5   & 0.26 $\pm$ 0.08 & 0.38 $\pm$ 0.10 & 0.12 $\pm$ 0.04 & 0.19 $\pm$ 0.05 & 0.32 $\pm$ 0.08 & 0.41 $\pm$ 0.06 & 0.27 $\pm$ 0.09 & 0.28 $\pm$ 0.07 \\
			0.94  & 0.50 $\pm$ 0.05 & 0.85 $\pm$ 0.15 & 0.14 $\pm$ 0.02 & 0.18 $\pm$ 0.04 & 0.39 $\pm$ 0.08 & 0.15 $\pm$ 0.03 & 0.30 $\pm$ 0.07 & 0.30 $\pm$ 0.08 \\
			\midrule
			\textbf{Mean} & \textbf{0.46 $\pm$ 0.050} & 0.64 $\pm$ 0.065 & 0.13 $\pm$ 0.014 & 0.14 $\pm$ 0.020 & 0.30 $\pm$ 0.033 & \textbf{0.23 $\pm$ 0.022} & 0.24 $\pm$ 0.031 & 0.25 $\pm$ 0.031 \\
			\bottomrule
		\end{tabular}
	}
\end{table*}

\begin{figure}[t!]
    \centering
    \includegraphics[width=\linewidth]{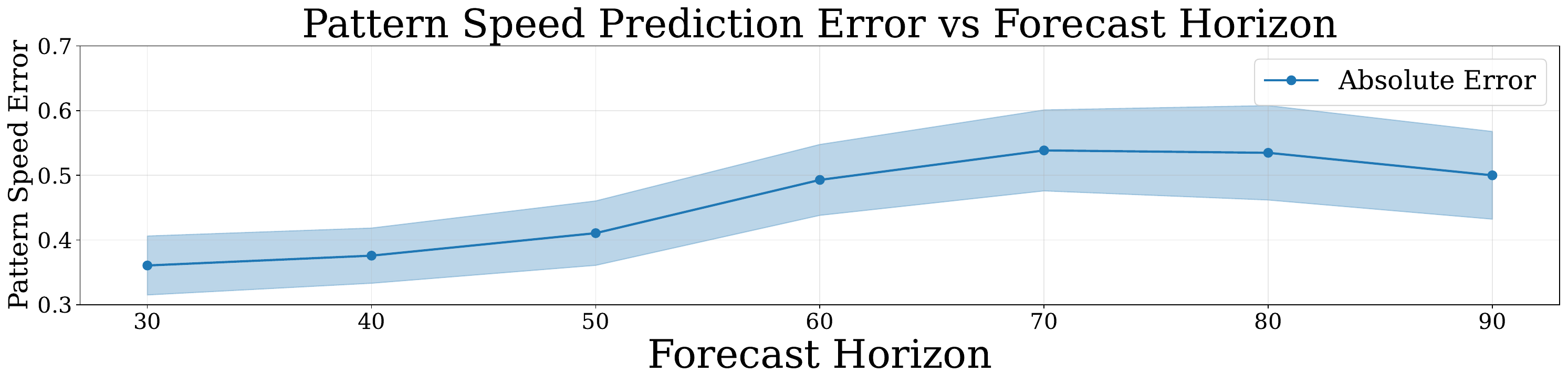}
    \caption{\textbf{Pattern Speed Error Plateaus vs. Horizon} }
    \label{fig:horizon_ablation}
\end{figure}

\paragraph{Feature Estimation Stability} 
In all our plasma feature estimate, we select a fixed forecast horizon of 60 frames (300 \M). To justify this choice, we conduct an ablation study on the forecast horizon from 30 frames to 90 frames. Note that beyond the lower bound, the measured feature becomes subject to noise and thus not meaningful, while beyond the upper bound the forecast becomes entirely uncorrelated. \cref{fig:horizon_ablation} shows a regime where pattern speed error \textbf{plateaus} ($\approx$ 60--80 frames), indicating that feature estimates remain stable even as frames decorrelate. The 60-frame horizon is a sweet spot to minimize chaotic drift with enough frames to avoid noisy measurements. The ablation indicates that the \codename pipeline forecasts roll-outs that match \textbf{spatio-temporal} statistics of GRMHD dynamics, even if pixel-level trajectory tracking becomes impossible due to chaos. 

\begin{figure}[b]
	\centering
	\includegraphics[width=\linewidth]{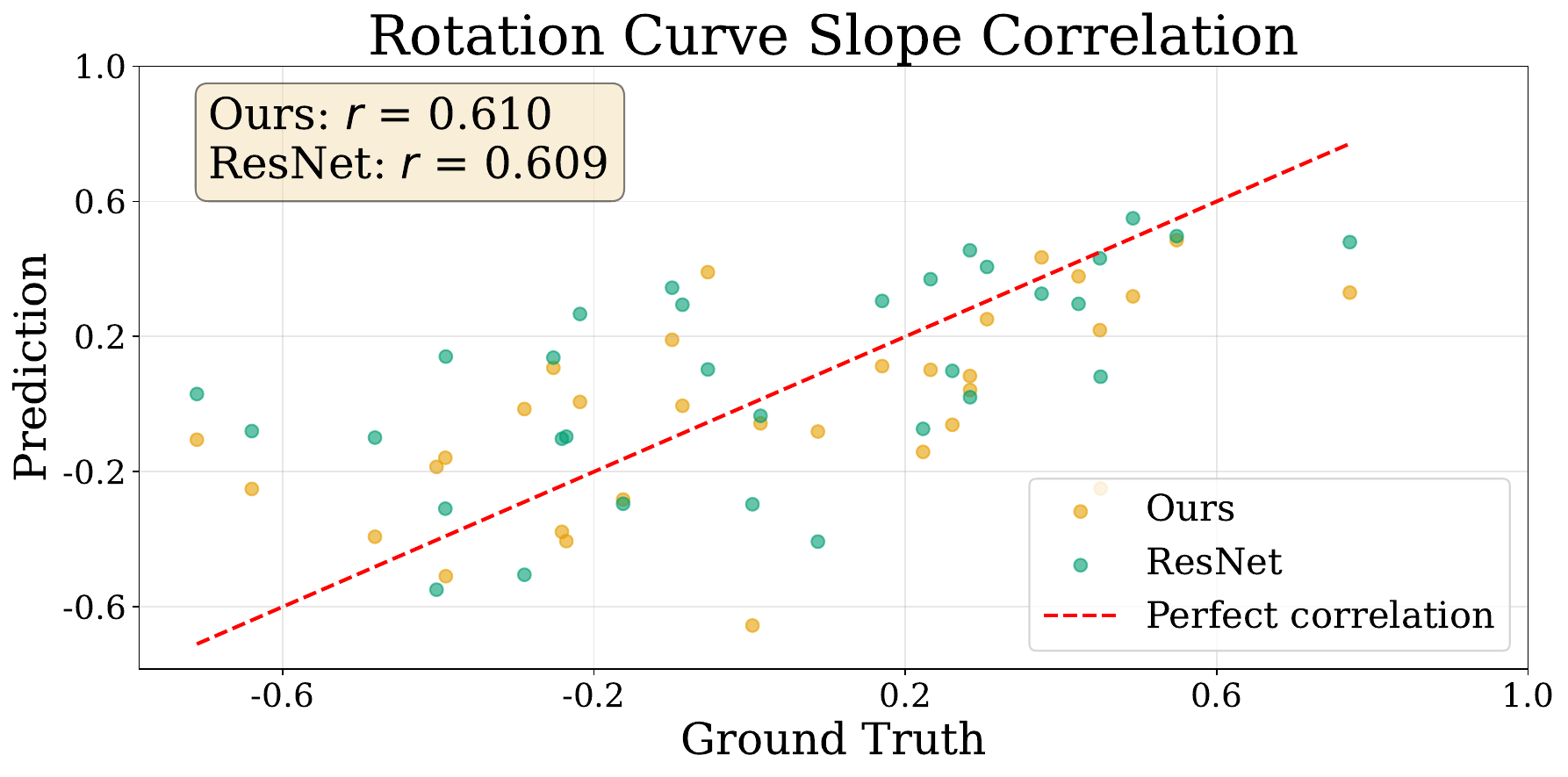} \\
	\caption{\textbf{Rotation Curve Slope Correlation:} Two methods show comparable correlation.
	}
	\label{fig:slope_correlation}
\end{figure}

\begin{figure}[h]
	\centering
	\includegraphics[width=\linewidth]{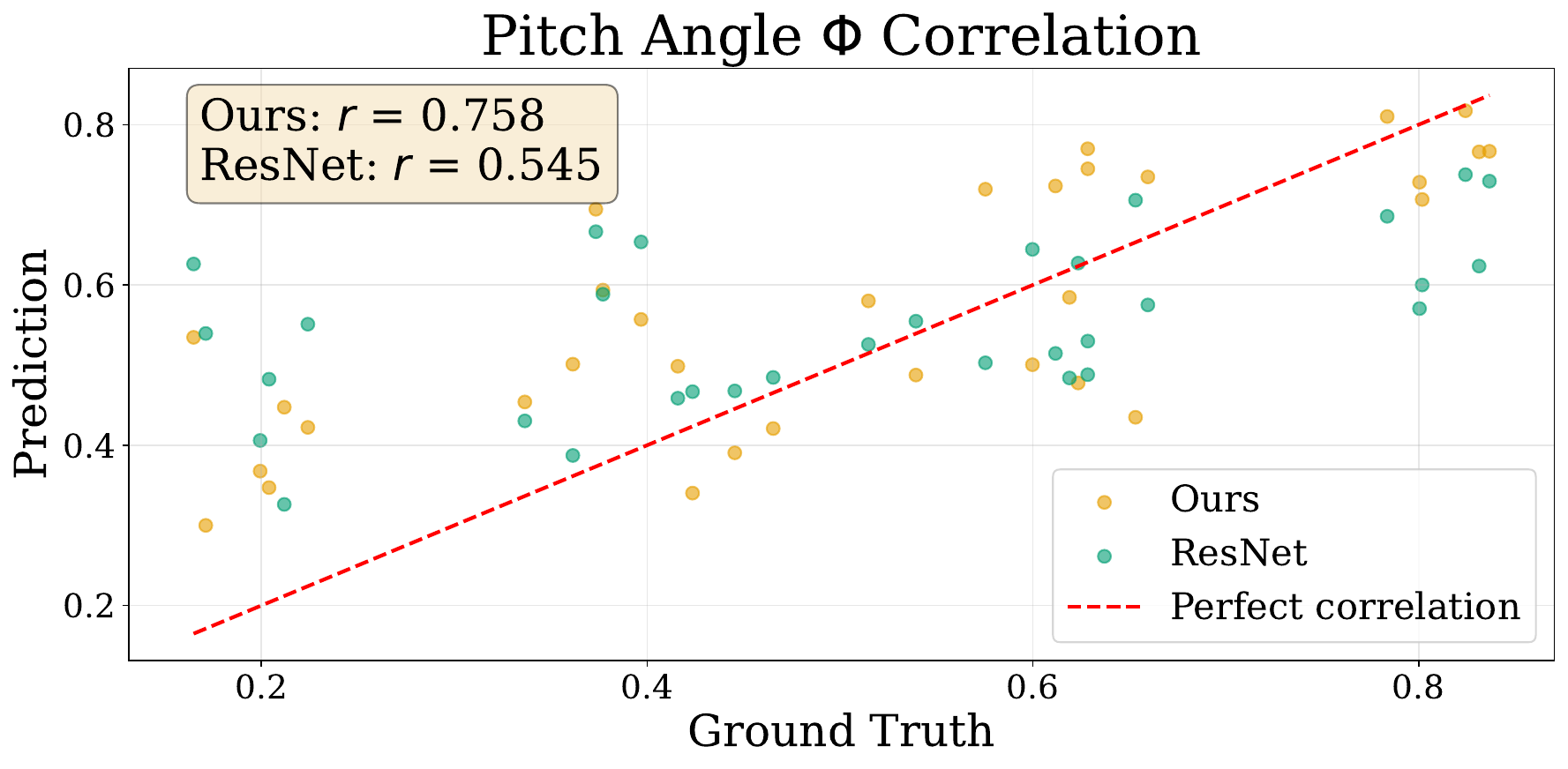} \\
	\caption{\textbf{Pitch Angle Correlation:} \codename demonstrates a stronger correlation.
	}
	\label{fig:pa_correlation}
\end{figure}

\begin{figure}[h]
	\centering
	\includegraphics[width=\linewidth]{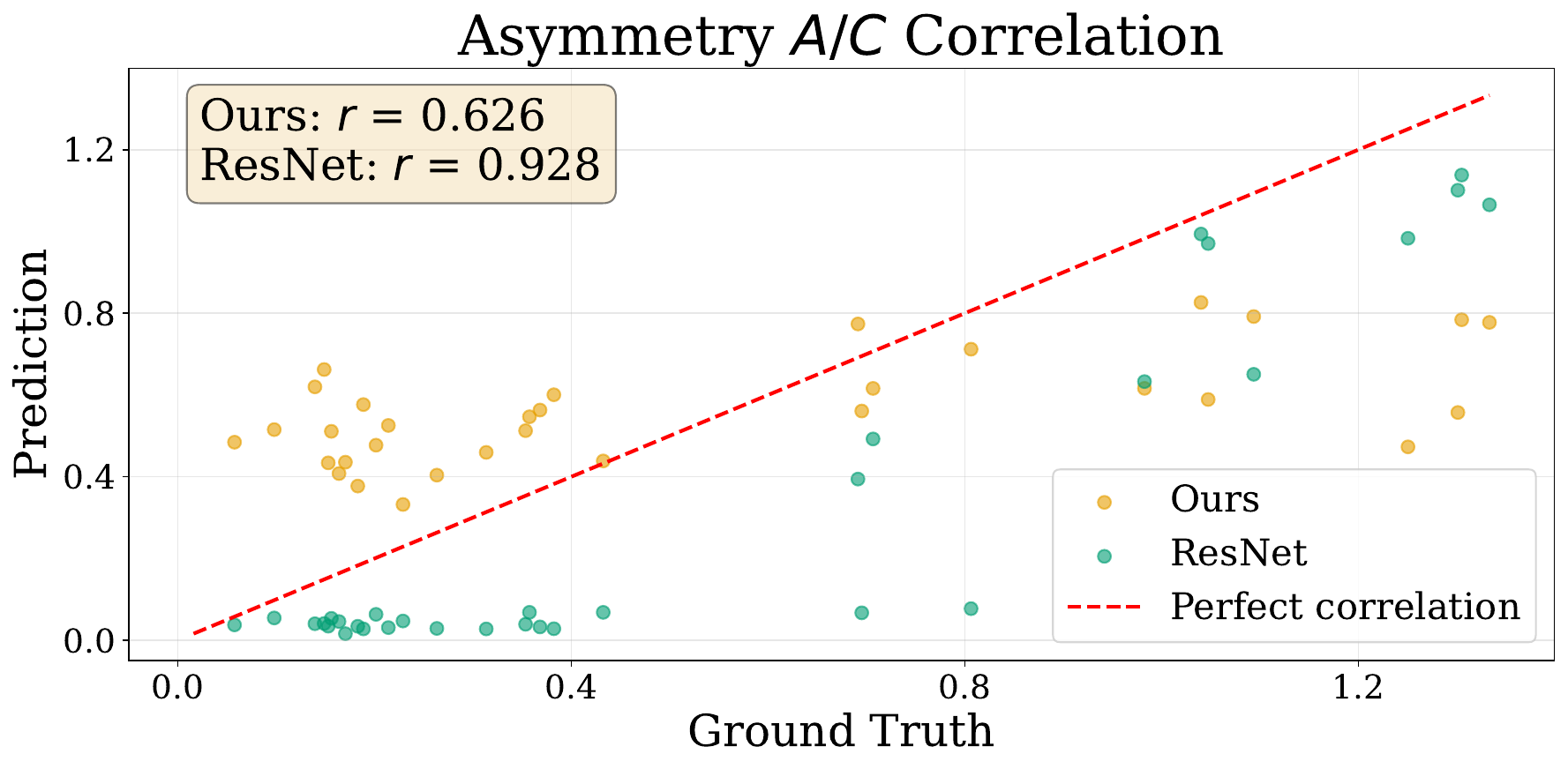} \\
	\caption{\textbf{Asymmetry Correlation:} although ResNet has a stronger correlation, its predictions are highly biased towards 0 when asymmetry is low.
	}
	\label{fig:asym_correlation}
\end{figure}

\paragraph{Feature Correlation Results} \vspace{-5pt}
To complement \cref{fig:feature_correlation}, we present the correlation plots for the three other features. 
\cref{fig:slope_correlation} shows comparable correlation between our dynamics-based approach and the ResNet baseline. \cref{fig:pa_correlation} demonstrates a stronger correlation for \codename. Finally, \cref{fig:asym_correlation} shows a better correlation for ResNet, but most of ResNet's predictions are biased towards zero. This is shown in the cluster of ResNet predictions on the bottom left. 

\subsection{Physics Inference Model}
\subsubsection{Robustness Study}
\cref{tab:parameter_accuracy} demonstrates \codename's robustness to different blur levels of the input image, which is a key advantage of our pipeline compared to a supervised baseline. We study the performance of physics inference with additional noise scenarios: (1). a salt-and-pepper noise is injected, randomly setting 1\% of the pixels to min or max values with equal probability; (2). the frame is horizontally shifted by 5\% of its width, while the lost pixels are filled with zeroes. In both scenarios, \cref{tab:robustness} shows \codename beating the ResNet baseline in black hole inclination and spin classification. Note that our pipeline demonstrates exceptional robustness in spin classification, as the accuracy barely degrades at all. 

\begin{table}[h]
	\centering
	\caption{Robustness analysis of \codename versus the ResNet baseline under input noise scenarios.  \codename demonstrates stability in spin accuracy under noise and translation.}
	\label{tab:robustness}
	\resizebox{\columnwidth}{!}{%
		\begin{tabular}{llcc}
			\toprule
			\textbf{Scenario} & \textbf{Model} & \textbf{Inclination Acc. } & \textbf{Spin Acc. } \\
			\midrule
			\multirow{2}{*}{\textbf{Only Blurring}} & ResNet50 & 47.19 & 67.66 \\
			& \textbf{\codename} & \textbf{56.41} & \textbf{69.22} \\
			\midrule
			Blurring & ResNet50 & 33.12\scriptsize{\textcolor{red}{(-14.07)}} & 52.19 \scriptsize{\textcolor{red}{(-15.47)}} \\
			\textbf{+ Salt \& Pepper (1\%)}& \textbf{\codename} & \textbf{45.16} \scriptsize{\textcolor{red}{(-11.25)}} & \textbf{67.50} \scriptsize{\textcolor{red}{(-1.72)}} \\
			\midrule
		    Blurring & ResNet50 & 28.75 \scriptsize{\textcolor{red}{(-18.44)}} & 53.91 \scriptsize{\textcolor{red}{(-13.75)}} \\
			\textbf{+ Translation (5\%)}& \textbf{\codename} & \textbf{33.44} \scriptsize{\textcolor{red}{(-22.97)}} & \textbf{73.44} \scriptsize{\textcolor{teal}{(+4.22)}} \\
			\bottomrule
		\end{tabular}%
	}
\end{table}

\subsubsection{Interpretability Analysis}

\paragraph{XGBoost Importance Scores} 
Gradient Boosting Trees provide direct insight on the importance of each feature in classification. \cref{tab:feature_importance} lists normalized importance scores globally, for spin classification, and for inclination classification. 

\begin{table}[h]
	\centering
	\caption{XGBoost Feature Importance Scores. The \textbf{Global} column represents the importance for the joint (Spin, Inclination) classification task. The specific \textbf{Inclination} and \textbf{Spin} columns show which features dominate when predicting either of them. Notably, \textbf{Pattern Speed} is the primary feature for Inclination inference, while \textbf{asymmetry} is the  critical feature for Spin.}
	\label{tab:feature_importance}
	\resizebox{\columnwidth}{!}{%
		\begin{tabular}{lccc}
			\toprule
			\textbf{Plasma Feature} & \textbf{Global (Joint)} & \textbf{Inclination} & \textbf{Spin} \\
			\midrule
			Pattern Speed $(\Omega_p)$ & 0.2625 & \textbf{0.3683} & 0.2314 \\
			Pitch Angle $(\Phi)$ & 0.2686 & 0.2065 & 0.2567 \\
			Asymmetry & \textbf{0.3314} & 0.2809 & \textbf{0.3833} \\
			Rot. Curve Slope & 0.1376 & 0.1443 & 0.1285 \\
			\bottomrule
		\end{tabular}%
	}
\end{table}

\paragraph{Shapley Additive Explanations (SHAP)} \vspace{-5pt}
SHAP uses shapley values from game theory to explain the output of machine learning models, and it is frequently used for tree interpretability ~\cite{lundberg2017unified}. SHAP informs us how and in which direction a feature influences a particular classification result, such as positive/negative spin or edge-on/face-on inclination. 

\begin{figure}[b]
	\centering
    \includegraphics[width=\linewidth]{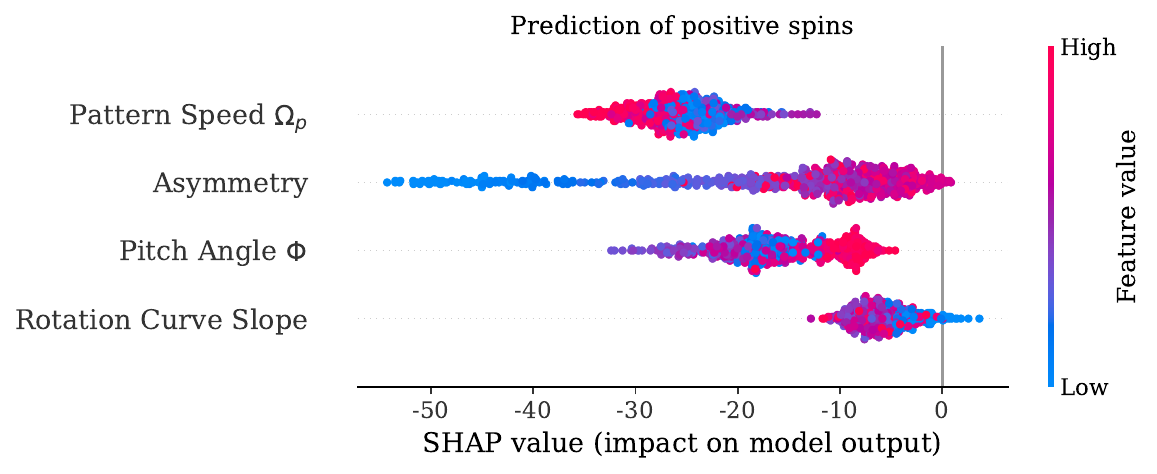} \\
	\includegraphics[width=\linewidth]{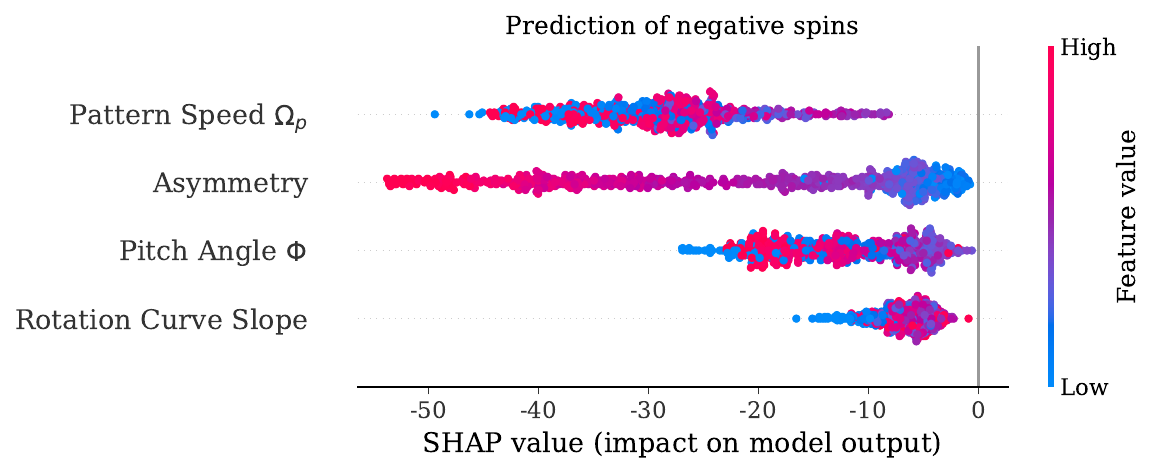} \\
	\caption{\textbf{SHAP analysis for spin classification:} SHAP values indicate asymmetry acts as the main feature for determining spin. High asymmetry values (red dots) push the models towards predicting positive spin (top plot), while low asymmetry values (blue dots) are strongly associated with negative spin predictions (bottom plot). 
	}
	\label{fig:shap_spin}
\end{figure}

\begin{figure}[h]
	\centering
    \includegraphics[width=\linewidth]{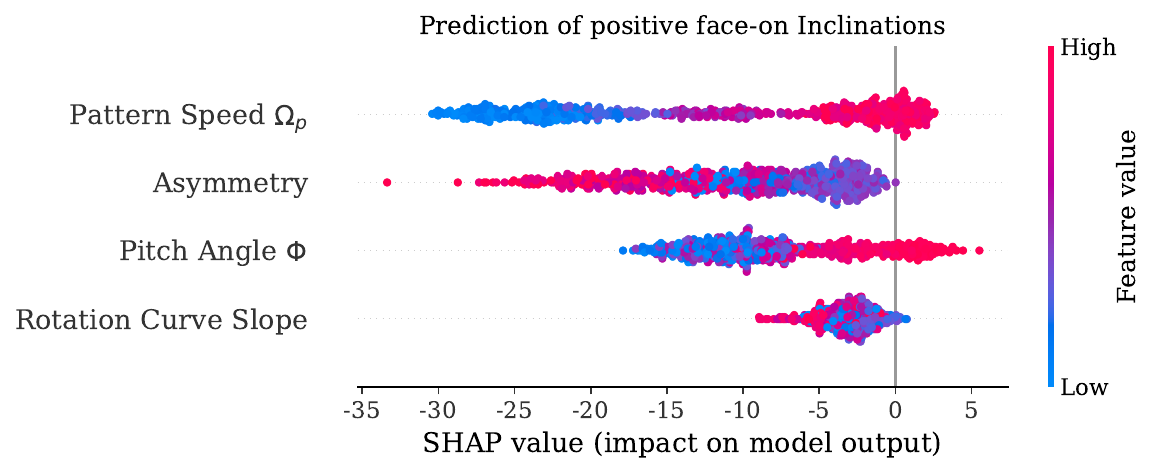} \\
	\includegraphics[width=\linewidth]{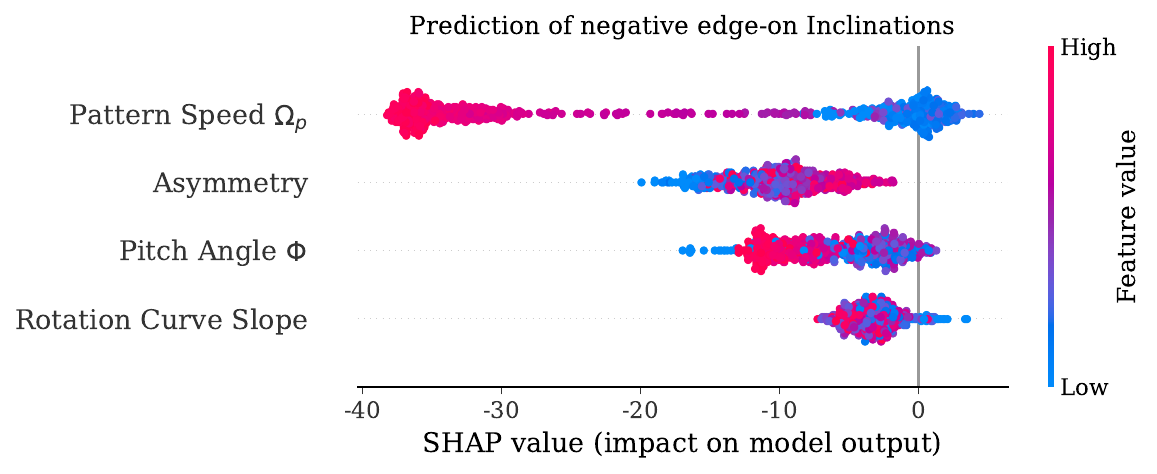} \\
	\caption{\textbf{SHAP analysis for inclination classification:} 
    Different from spin, pattern speed is the main feature for determining inclination. High pattern speed values (red dots) indicate positive inclinations, and vice versa. Notably, high pitch angle values contribute to face-on classifications, whereas low rotation curve slopes contribute to edge-on.   
	}
	\label{fig:shap_inc}
\end{figure}

\cref{fig:shap_spin} and \cref{fig:shap_inc} show two interpretability studies on spin and inclination classification. The results are consistent with prior astrophysics research: (1) asymmetry is an important observable that correlate with spin $a_*$; (2) pattern speed is consistent with the sign of $\cos(inclination)$. These imply that our XGBoost physics inference model is \textit{right for the right reasons}. Interestingly, our results also reveal unexpected relationships, such as large pitch angle contributing to face-on inclination classification and low rotation curve slopes contributing to edge-on.

\begin{table}[t]
\centering
\small
\caption{\textbf{M87* EHT eval via test-time augmentation (TTA).}}
\vspace{-5pt}
\begin{tabular}{lcc}
\hline
\textbf{Augmentation} & \textbf{Count} & \textbf{Correct Rotation (\%)}  \\
\hline
Base (original) & 5  & 4/5 (80.0)   \\
Translation ($\Delta\!\in\!\{-2,-1,1\}$) & 15 & 10/15 (66.7) \\
Blur (PSF scale $\times\{0.9,1.1\}$) & 10 & 7/10 (70.0)  \\
Correlated noise (corr=1.0) & 20 & 14/20 (70.0)  \\
\hline
All TTA (perturbed only) & 45 & 31/45 (68.9) \\
\hline
\end{tabular}
\label{tab:m87_tta}
\vspace{-5pt}
\end{table}

\begin{figure}[t!]
    \centering
    \includegraphics[width=\linewidth]{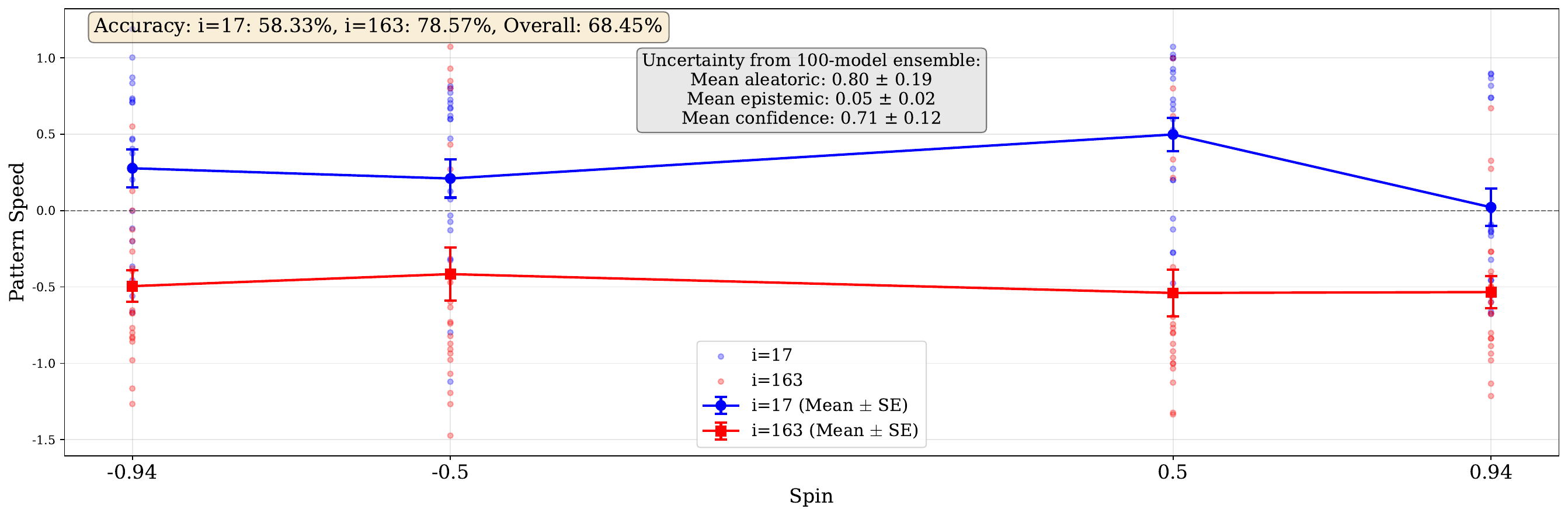}
    \caption{\textbf{M87* Pattern Speed Uncertainty Quantification.} }
    \label{fig:m87_ps_dist}
    \vspace{-5pt}
\end{figure}

\subsection{Extrapolation to M87*}
To complement qualitative results in \cref{sec:m87_results}, we prioritize real-data evaluation by expanding to \textbf{5 M87* EHT image reconstructions}. \codename measures pattern speeds on roll-outs from EHT images and matches the expected rotation on 4/5 real reconstructions (\cref{tab:m87_tta}). To quantify robustness, we apply Test-Time Augmentation via image perturbations, yielding 68.9\% rotation matching. Furthermore, for uncertainty quantification on M87*, we evaluate \codename on 168 simulation frames using a 100-model ensemble (\cref{fig:m87_ps_dist}). The model achieves 68.5\% zero-shot accuracy with high confidence 0.71$\pm$0.12 and low epistemic uncertainty 0.05.

\section{Supplement on GRMHD and EHT Imaging}
\label{app:background}
\subsection{EHT Imaging}
We provide further explanation on sources contributing to the blurriness of EHT images. Chief among these is interstellar scattering, which causes refractive and diffractive blurring of the incoming wavefronts ~\cite{Johnson_2018, Issaoun_2021}. In addition, the sparse coverage of very-long-baseline interferometry (VLBI) leads to incomplete sampling of Fourier modes of the source brightness distribution, limiting image fidelity and contributing to intrinsic maximum resolution of the observations. Finally, the finite aperture synthesis and temporal averaging required to build an image further smear small-scale variability ~\cite{sgr_a_paper1, sgr_a_paper3}. 

Now, we expand upon quantities that characterize black hole dynamics, pattern speed and pitch angle ~\cite{baubock_spiral_prep, conroy2023rotation}. A small opening angle corresponds to tightly wound spirals, whereas a large opening angle indicates more open, loosely wound structures. This metric is sensitive to magnetized turbulence and the development of instabilities in the accretion flow, and has been used in GRMHD analyses as a diagnostic of dynamical state ~\cite{Ricarte_2022_pitch_angle}. In EHT applications, both the pattern speed and spiral opening angle can in principle be inferred from the temporal variability of interferometric observables, but in practice the sparse telescope coverage and scattering make these classical approaches extremely challenging. These limitations motivate the development of dynamics-aware inference frameworks, such as the one we introduce, which aim to extract these quantities directly from data or surrogate-generated movies in a robust manner.

\subsection{GRMHD Simulations} 
In this subsection, we provide information on the physical model and scientific codebase underlying the GRMHD simulations.
GRMHDs model the dynamic evolution of the accretion flow through evolving the magnetized relativistic fluid by solving the source-free evolution equation for the magnetic field, constrained by the no-monopole condition, equations of particle number conservation and conservation of energy-momentum ~\cite{Porth_2019}. 

Typically, GRMHDs are evolved for times longer than $3 \times 10^4 \,\, GM / c^3.$ GRMHD accretion-flow models are typically categorized into two regimes: \textit{Standard and Normal Evolution} (SANE) and \textit{Magnetically Arrested Disks} (MAD)  ~\cite{Bisnovatyi_MAD_1976, Igumenshchev_MAD_2003, Narayan_MAD_2003, Tchekhovskoy_MAD_2011}. \textbf{We limit our dataset to MAD simulations} as they are currently favored for explaining horizon-scale behavior in both M87* and Sgr~A* \cite{m87_paper5, sgrA_paper_v}. In this regime, strong magnetic flux accumulates near the black hole, compressing the accretion disk into a compact, highly dynamic structure. The GRMHD models used in this paper are from the ``Illinois v3'' library. The fluid simulations are generated using the KHARMA code and imaged with the {\tt ipol} code ~\cite{kharma_prather_2024,ipol_paper}.

Although the same GRMHD framework applies to both M87* and Sgr A*, their observational contexts differ in ways that motivate our approach. M87*, with a mass of $\sim 6.5 \times 10^9$ in solar mass units, evolves on month-long dynamical timescales, making time-averaged simulation images appropriate for comparison with EHT data ~\cite{m87_paper1, eht2023_persistent_shadow}. In contrast, Sgr A* is $\sim 10^3$ times less massive and varies on minute timescales within a single observing run, where time-averaging washes out physical structure ~\cite{sgr_a_paper1, sgr_a_paper4}. This rapid variability highlights the need for dynamics-aware methods, such as the framework we develop, that can model and interpret the evolving structure of the accretion flow.

\end{document}